\pdfoutput=1

\documentclass[11pt]{article}

\usepackage{acl}

\usepackage{times}
\usepackage{latexsym}

\usepackage[T1]{fontenc}

\usepackage[utf8]{inputenc}

\usepackage{microtype}

\usepackage{inconsolata}

\usepackage{amsthm,amsmath,amssymb}
\usepackage{mathrsfs}
\usepackage{float}  
\usepackage{subcaption}
\usepackage{graphicx}
\usepackage{multirow}
\usepackage{booktabs}
\usepackage{algorithm, algorithmic}
\usepackage{stfloats}
\PassOptionsToPackage{colorlinks,linkcolor=blue}{hyperref}
\newtheorem{definition}{Definition}[section]
\newtheorem{hypothesis}{Hypothesis}

\title{Focus on Your Question! Interpreting and Mitigating Toxic CoT \\ Problems 
in Commonsense Reasoning}

\author{Jiachun Li\textsuperscript{1,2}, Pengfei Cao\textsuperscript{1,2}, Chenhao Wang\textsuperscript{1,2},  Zhuoran Jin\textsuperscript{1,2}, Yubo Chen\textsuperscript{1,2,}\footnotemark[1] \\ \textbf{Daojian Zeng\textsuperscript{3}, Kang Liu\textsuperscript{1,2}, Jun Zhao\textsuperscript{1,2,}\footnotemark[1]} \\ \textsuperscript{1}School of Artificial Intelligence, University of Chinese Academy of Sciences \\ \textsuperscript{2}The Laboratory of Cognition and Decision Intelligence for Complex Systems, \\ Institute of Automation, Chinese Academy of Sciences  \\ \textsuperscript{3}Hunan Normal University \\
        \footnotesize{\texttt{\{jiachun.li, pengfei.cao, chenhao.wang, zhuoran.jin, yubo.chen, kliu, jzhao\} @nlpr.ia.ac.cn }} \\ \footnotesize{\texttt{zengdj916@163.com}} }
\begin{document}

\maketitle
\renewcommand{\thefootnote}{\fnsymbol{footnote}}
\footnotetext[1]{Corresponding authors.}
\renewcommand{\thefootnote}{\arabic{footnote}}
\begin{abstract}
Large language models exhibit high-level commonsense reasoning abilities, especially with enhancement methods like Chain-of-Thought (CoT). However, we find these CoT-like methods lead to a considerable number of originally correct answers turning wrong, which we define as the Toxic CoT problem. To interpret and mitigate this problem, we first utilize attribution tracing and causal tracing methods to probe the internal working mechanism of the LLM during CoT reasoning. Through comparisons, we prove that the model exhibits information loss from the question in the shallow attention layers when generating rationales or answers. Based on the probing results, we design a novel method called $\mathbb{RIDERS}$ (\textbf{R}esidual decod\textbf{I}ng and s\textbf{ER}ial-position \textbf{S}wap), which compensates for the information deficit in the model from both decoding and serial-position perspectives. Through extensive experiments on multiple commonsense reasoning benchmarks, we validate that this method not only significantly eliminates Toxic CoT problems (decreased by \textbf{23.6\%}), but also effectively improves the model's overall commonsense reasoning performance (increased by \textbf{5.5\%}).
\end{abstract}

\section{Introduction}
With the increase in scale, large language models (LLMs) have demonstrated outstanding performance in different tasks \citep{evalmodel_di,chinamodel_di,moss_mir,conflict_zhuoran}, among them, commonsense reasoning has received significant attention due to its importance for general intelligence \citep{cnatomic,commonse_reasoning_wch,commonsense_mir}. In this task, researchers have proposed a series of chain-of-thought (CoT) like techniques to elicit models' potential abilities (e.g. Self-Consistency \citep{sc}, Least-to-Most \citep{l2m}, Reflexion \citep{reflexion}). Through them, LLMs can generate reasonable rationales and improve their reasoning performance.

\begin{figure}[tbp]
    \centering
    \includegraphics[width=0.48\textwidth]{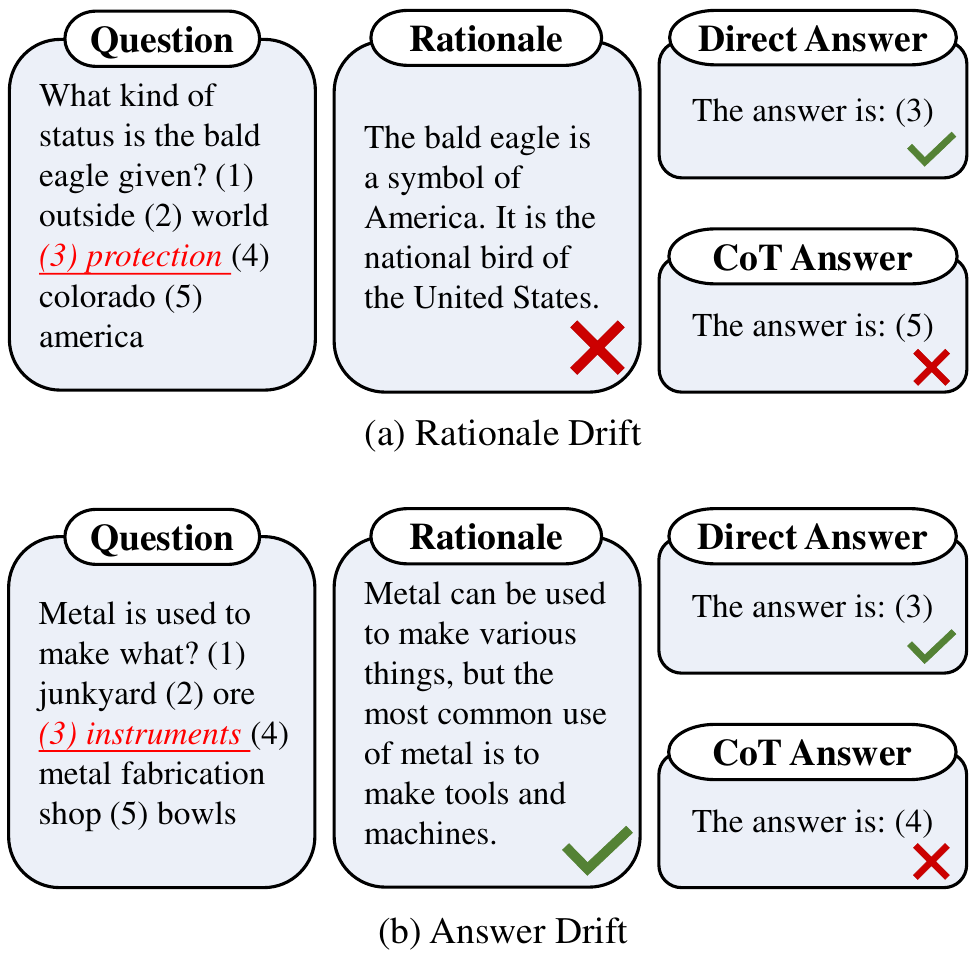}
    \caption{Two examples for the Toxic CoT problem.}
    \label{fig:example}
\end{figure}

While these works have made great progress, we notice an overlooked problem in them, which we define as \textbf{Toxic CoT} --- Sometimes LLMs can directly provide correct answers to questions, but after applying CoT-like methods, it brings extra reasoning paths to models, causing their answers to be wrong.\footnote{Notably, our definition here differs from the toxicity in generated content \citep{toxicity}, emphasizing the harm that CoT can bring to the model.} Figure \ref{fig:example} illustrates two main error types of this problem --- Rationale Drift and Answer Drift. Specifically, for the Rationale Drift case, given the question ``\textit{What kind of status is the bald eagle given?}'', the model can directly give the correct answer ``\textit{protection}''. However, in the rationale, the model explains ``\textit{what is the bald eagle}'' as ``\textit{a symbol of America}'', which has a semantic drift from the question. Thus, the model chooses the wrong option ``\textit{america}'' based on the drifting rationale. For the Answer Drift case, given the question ``\textit{Metal is used to make what?}'', the model can directly answer ``\textit{instruments}''. It can also generate a correct rationale ``\textit{metal is to make tools and machines}'', but when answering based on the rationale, the model drifts from it and selects the incorrect option ``\textit{(4)}''. We further conduct a statistical analysis over extensive commonsense reasoning datasets and find that, among all CoT errors, this problem accounts for \textbf{37\%} for the white-box model and \textbf{33\%} for the black-box model on average, indicating this problem has become a crucial bottleneck in CoT reasoning.\footnote{Appendix \ref{sec:early-stat} presents the details of settings and results in this statistical experiment.}

So what is the mechanism behind this issue? In this paper, we attempt to answer this question by probing the inner workings of the LLM's CoT reasoning. Specifically, we first make initial observations on examples of Rationale Drift and Answer Drift issues, which suggest that the model likely misses some important information from the question when generating corresponding rationales or answers. To further verify these findings, we use attribution tracing and causal tracing methods to probe the LLM in two stages (rationale generation stage and answer generation stage). By employing these methods under various experimental settings, \textbf{we find that there is a significant loss of information flow from the question in the shallow attention layers when generating drifting rationales and answers.} Therefore, we interpret the Toxic CoT problem as the model lacking information from the question in the two stages.
 
To validate our interpretation and mitigate this problem, we design an approach called $\mathbb{RIDERS}$ (\textbf{R}esidual decod\textbf{I}ng and s\textbf{ER}ial-position \textbf{S}wap) based on the interpretation. Concretely, for the Rationale Drift issue, we devise a decoding algorithm, promoting the model to generate tokens that pay more attention to question contexts. For the Answer Drift issue, we swap the positions of the output sequence, reducing the information loss from the question to the final prediction. We evaluate our method on five commonsense reasoning benchmarks and conduct extensive experiments. The results not only prove our interpretation, but also indicate that our method is effective in addressing the Toxic CoT problem and improving the model's overall commonsense reasoning abilities. 

We summarize the contribution of this paper as follows: 

(1) We identify a crucial bottleneck affecting LLM's reasoning performance called the Toxic CoT problem, probe this issue through attribution tracing and causal tracing methods, and interpret the mechanism behind it as the model missing information from questions in shallow attention layers. The results contribute to a more in-depth understanding of the LLM's reasoning mechanisms.

(2) To mitigate the Toxic CoT problem, we introduce $\mathbb{RIDERS}$, which effectively compensates for the internal information loss during CoT reasoning from decoding and serial-position perspectives. 

(3) We conduct extensive experiments on various benchmarks. The results not only verify the rationality of our interpretation, but also demonstrate the effectiveness of our method in addressing the Toxic CoT problem (the proportion of the problem decreased by \textbf{23.6\%}) and enhancing commonsense reasoning performance (overall accuracy increased by \textbf{5.5\%}). Our code is available at: 
\href{https://github.com/BugMakerzzz/toxic_cot}{https://github.com/BugMakerzzz/toxic\_cot}.

\section{Problem Statement} \label{sec:2}
\subsection{Toxic Chain of Thought Reasoning}\label{sec:2.1}
We start our work by formally defining the Toxic CoT problem as follows: \footnote{In practice, CoT-type methods all have Toxic CoT problems, but to simplify the work, this paper mainly focuses on the basic CoT prompting.}
\begin{definition}[\textbf{Toxic CoT}]
Given a question $q$ and the correct answer $o^*$, if the model's output $\mathcal{M}$ meets the following conditions, it is considered a case of Toxic CoT:
\begin{equation*}
   \footnotesize{
   \begin{aligned}
    o^* &= \mathcal{M}(q, P_d) \wedge   o^*\neq \mathcal{M}(q, P_c)\\ 
    \end{aligned}} 
\end{equation*}
where $o^* = \mathcal{M}(q, P_d)$ indicates the model's direct answering for $q$ is correct, $o^*\neq \mathcal{M}(q, P_c)$ indicates the model's cot-like answering for $q$ is wrong, $P_d, P_c$ are the corresponding prompts. 

\end{definition} 

\subsection{Two-stage Drift Issues}\label{sec:2.2}
To investigate the reasons for the problem, we classify these Toxic CoT cases and identify a main error causing this problem (On average, it accounts for 67\% on two datasets, see more details in Appendix \ref{sec:Toxic_reason}). Furthermore, if we divide the CoT process into two stages: \textbf{rationale generation} and \textbf{answer generation}, there exist two types of issues in this error:
\begin{definition}[\textbf{Rationale Drift}] 
If the reasoning chain is factually correct but logically inconsistent with the question, this case is called ``Rationale Drift'' (see Figure \ref{fig:example}a).
\end{definition}
\begin{definition}[\textbf{Answer Drift}]
If the reasoning chain is both factually correct and logically consistent with the question, but the final answer is inconsistent with the rationale, this case is called ``Answer Drift'' (see Figure \ref{fig:example}b).
\end{definition}

\subsection{Hypothesis Formulation}\label{sec:2.3}
To provide a direction for subsequent probing experiments, here we attempt to propose hypotheses for the mechanism of issues by analyzing some examples.
For the Rationale Drift issue, the model tends to focus on part of the essential reasoning conditions in the question context. As an example, in Figure \ref{fig:example}a, the CoT only focuses on the ``\textit{bald eagle}'' in the question but misses another key information ``\textit{status}''. As for the Answer Drift issue, the model seems to be disrupted by CoT, losing attention to the question and resulting in an off-topic prediction. For instance, in Figure \ref{fig:example}b, though the CoT gives correct information ``\textit{to make tools and machines}'', the model can only predict the wrong answer ``\textit{(4) metal fabrication shop}''. This is likely because the model loses the question's target ``\textit{to make what}'' and directly copies the entity ``\textit{metal}'', which frequently appears in CoT, as the answer.
Therefore, we summarize our hypotheses as follows: 
\begin{hypothesis}\label{hypo1}
The Rationale Drift issue arises from the model lacking information from the question context in the rationale generation stage.
\end{hypothesis}
\begin{hypothesis}\label{hypo2}
The Answer Drift issue arises from the model lacking information from the question in the answer generation stage.
\end{hypothesis}
To validate the above hypotheses, in the following two sections, we conduct probing experiments, exploring the LLM's internal working mechanisms during the two stages of CoT reasoning.

\section{Tracing Information Flow in Rationale}\label{sec:3}
In this section, we aim to verify the Hypothesis \ref{hypo1} by tracing the information flow in the rationale generation stage. To this end, we start by describing our attribution tracing method ($\S$\ref{sec:3.1}). Through this method, we conduct comparative experiments between the correct reasoning and the drifting one, figuring out the mechanism behind the issue ($\S$\ref{sec:3.2}). At last, we use the attention score to validate our findings from another perspective ($\S$\ref{sec:3.3}).

\subsection{Tracing Method}\label{sec:3.1}
To investigate the roles of different model components in the rationale generation stage, we use attribution scores \citep{attribution, knowledgeneuron, labelworlds} to compute the contribution of a neuron $\omega$:
\begin{equation}\label{eq:attr}
   \footnotesize{
   \begin{aligned}
 Attr(\omega ) = \omega \odot \int_{\alpha=0}^{1} \frac{\partial F(\alpha \omega)}{\partial \omega} d \alpha  \approx \frac{\omega}{m} \odot \sum_{k=1}^{m} \frac{\partial F (\frac{k}{m} \omega )}{\partial \omega}
    \end{aligned}} 
\end{equation}
where $F(\cdot)$ represents the model's output. We compute the attribution score via Riemman approximation of the integration and $m$ is the number of approximation steps.
For neurons $A^{(l)}$ in the i-th attention layer, we sum the absolute values of scores on all attention heads to get the final attribution score. Since the attention module involves interactions between different tokens, we can compute the information flow between the question context $q$ and the CoT $c$ on it:
\begin{equation}\label{eq:infor_flow}
\footnotesize{\begin{aligned}
    &Q^{(l)}_{qc} = \frac{1}{|N|} \sum_{(i,j) \in C_{qc}} Attr(A^{(l)}_{i,j}) \\
    &C_{qc} = \{(i, j) |  q_s \leq i \leq q_e , \ c_s \leq j \leq c_e \}
    \end{aligned}}
\end{equation}
Here, $Attr(A^{(l)}_{i,j})$ represents the intensity of information flow from the $i$-th token to the $j$-th token in the $l$-th attention layer and $|N|$ denotes the number of CoT steps. More implementation details of this method are reported in Appendix \ref{sec:attr_trac}.

\begin{figure}[tbp] 
    \centering
    \begin{subfigure}[t]{.49\linewidth}
        \centering
	\includegraphics[width=\linewidth]{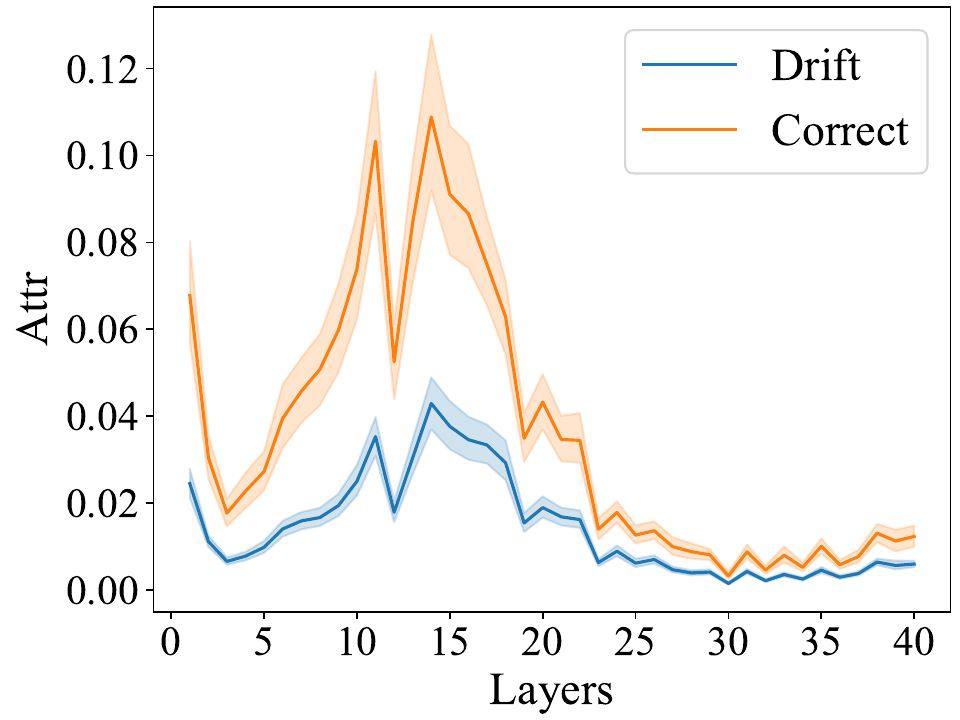}
        \caption{Llama2-13B}\label{fig:wino_attr_comp}
    \end{subfigure}
    \begin{subfigure}[t]{.49\linewidth}
        \centering
	\includegraphics[width=\linewidth]{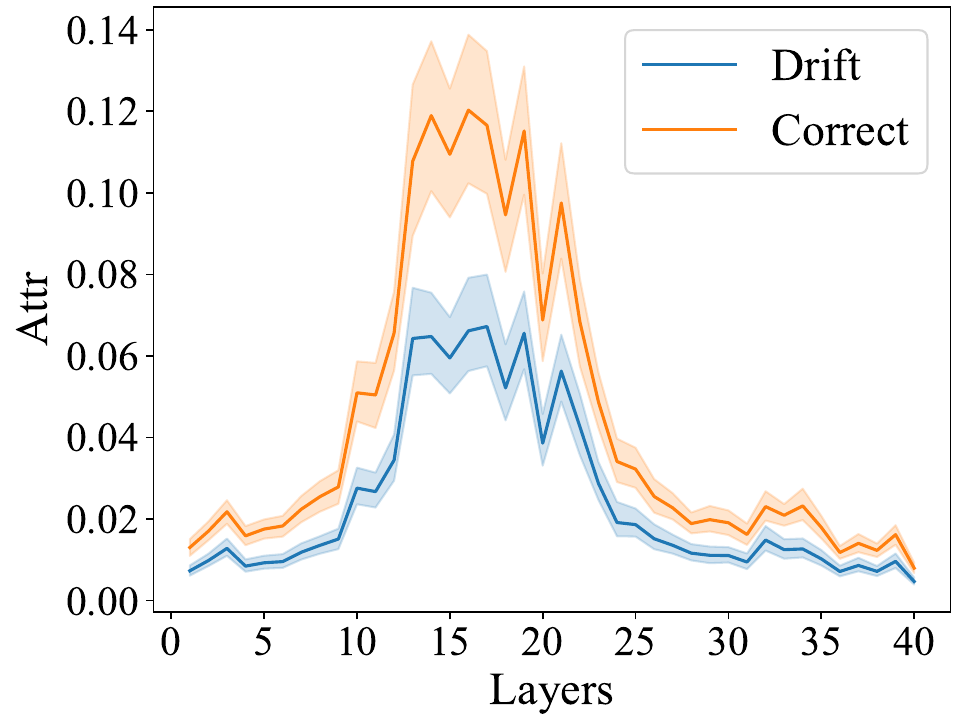}
        \caption{Baichuan2-13B}\label{fig:wino_attr_comp_b}
    \end{subfigure}
    \\
    \caption{Attribution tracing results on Winogrande.}
    \label{fig:attr_comp_wino}
\end{figure}

\subsection{Attribution Tracing Experiment}\label{sec:3.2}
\paragraph{Experimental Settings}
To validate the deficiency in Hypothesis \ref{hypo1}, we need to figure out the context information flow difference between generating a drifting rationale and a correct one. Thus, we first use golden labels as hints to generate correct CoTs in the drifting cases. Then, we compute the average information flow under these two cases and compare their results. We choose Llama2-13B \citep{llama2} and Baichuan2-13B \citep{baichuan2} as our probing models, since they are moderated-sized white-box models with decent CoT performance. For datasets, we select Winogrande \citep{winogrande} and CSQA \citep{csqa}.\footnote{Unless otherwise specified, we use the same models and datasets in the following probing experiments.} The detailed implementation of this experiment is shown in Appendix \ref{sec:attr_trac}.
\paragraph{Result and Analysis}
Figure \ref{fig:attr_comp_wino} illustrates our experimental results on Winogrande (The results on CSQA are shown in Appendix \ref{sec:attr_trac}). We can find that: \textbf{(Claim 1) When the Rationale Drift issue occurs, CoT receives less information from the question context compared to the correct case.} On both datasets and different models, there is a significantly lower information flow between the question context and CoT when the LLM generates a drifting rationale (the blue line) compared to the correct one (the orange line). This aligns with Hypothesis \ref{hypo1}. \textbf{(Claim 2) The shallow attention layers are crucial for LLMs to extract contextual information.} In all cases, both the information flow and the gap peak at around the 15th attention layer, indicating these layers are significant sites for the rationale generation. 

\begin{figure}[tbp] 
    \centering
    \begin{subfigure}[t]{.49\linewidth}
        \centering
	\includegraphics[width=\linewidth]{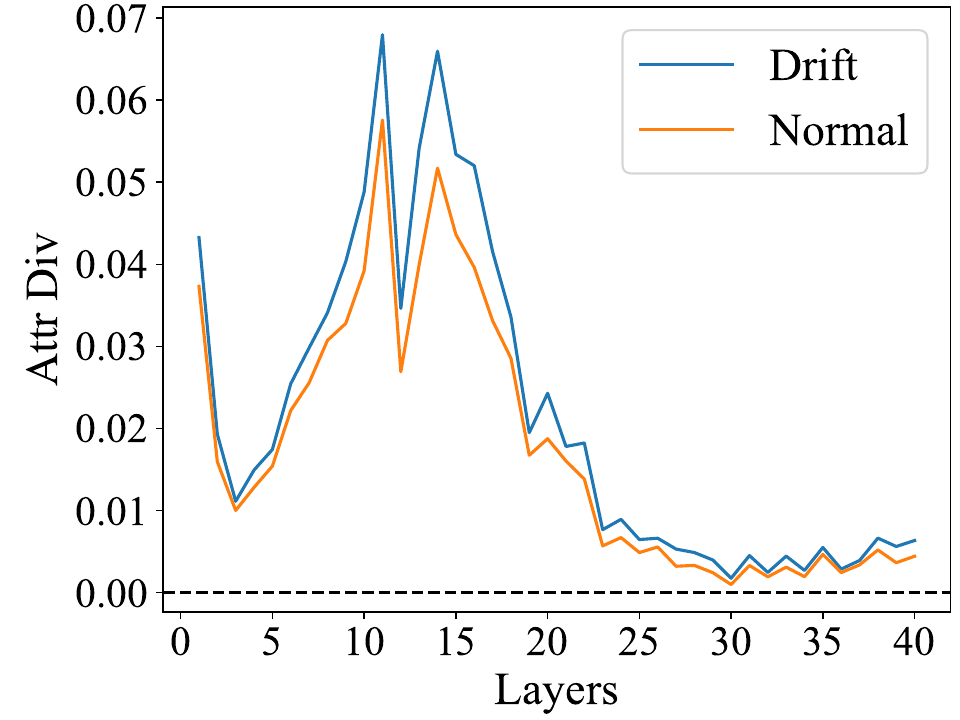}
        \caption{Llama2-13B}
    \end{subfigure}
    \begin{subfigure}[t]{.49\linewidth}
        \centering
	\includegraphics[width=\linewidth]{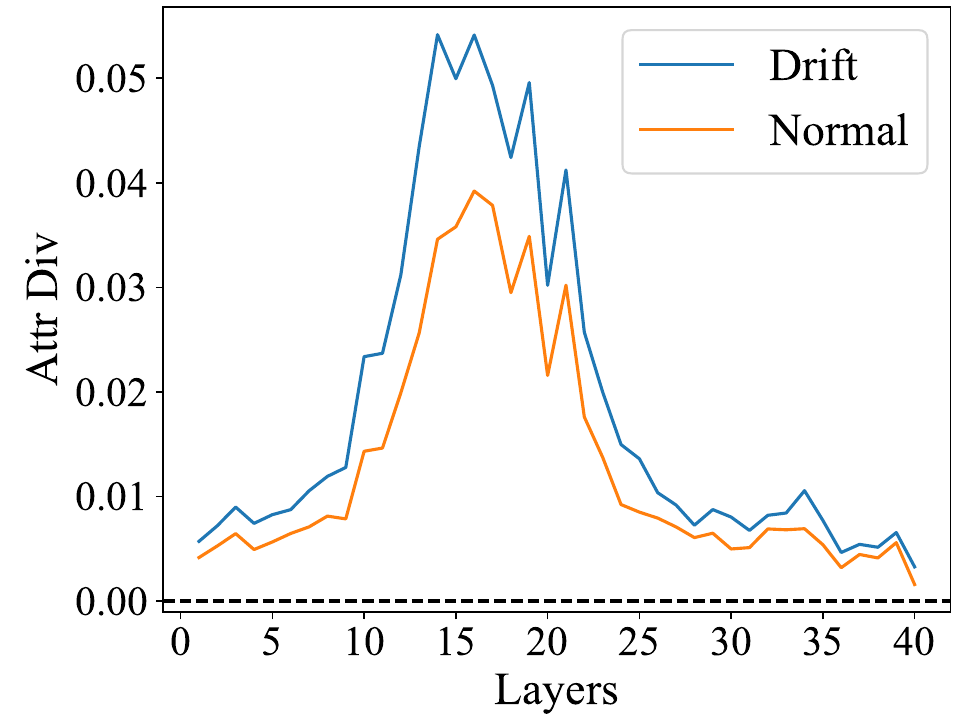}
        \caption{Baichuan2-13B}
    \end{subfigure}
    \\
    \caption{Information flow divergence comparison on Winogrande.}
    \label{fig:attr_comp_extra}
\end{figure}

\paragraph{Supplementary Experiment}
In the main experiment, we use golden labels to generate the correct CoT. To eliminate the influence of this additional factor on our results, we design a supplementary experiment. Concretely, we first use the label to generate CoTs from correct reasoning cases, which serves as a control group. Then, we compute the context information divergence between the newly generated CoT $c_n$ and the original one $c_o$, i.e.:
\begin{equation}\label{eq:attr_diff_score}
\footnotesize{
    \begin{aligned}
    Attr(c_n | c_o) = Q^{(l)}_{qc_n} - Q^{(l)}_{qc_o}
    \end{aligned}}
\end{equation}
where $q$ is the question context. We compare this divergence between the control and drifting group, whose results are reported in Figure \ref{fig:attr_comp_extra} and Figure \ref{fig:attr_comp_csqa_extra_b}. As we can see, the gap between the correct CoT and the drifting one (the blue line) is larger than the control group (the orange line) and the max divergence occurs in shallow layers (around the 15th layer). \textbf{This indicates that a correct CoT indeed gets more information flow from the context in shallow attention layers, validating the effectiveness of Claim 1 and 2.}

\begin{figure}[tbp] 
    \centering
    \begin{subfigure}[t]{.49\linewidth}
        \centering
	\includegraphics[width=\linewidth, trim=20 0 0 0, clip]{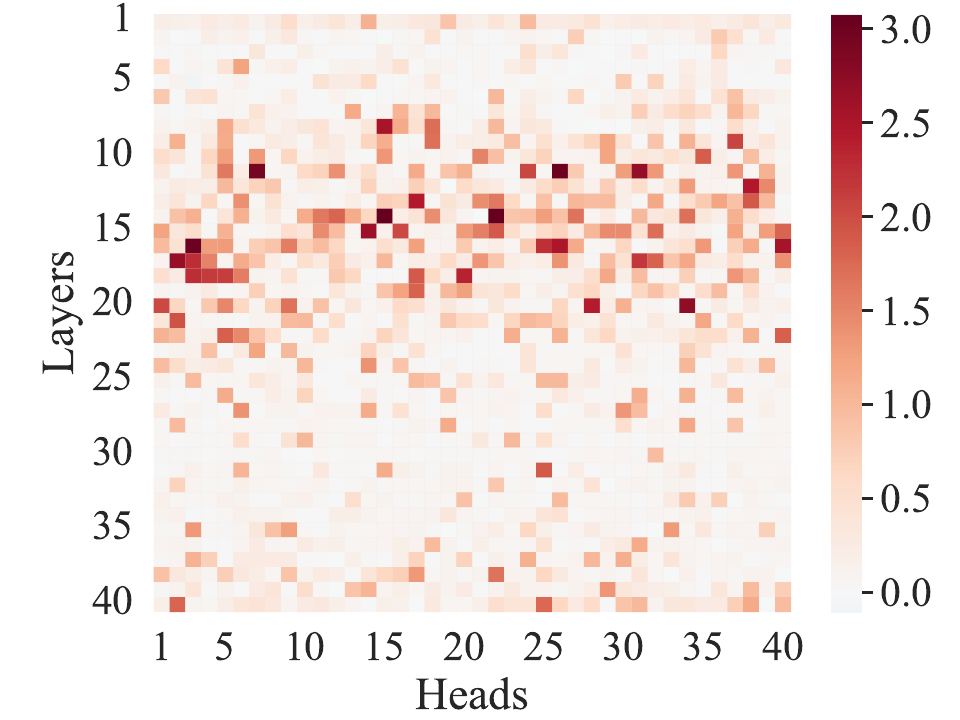}
        \caption{Winogrande}
    \end{subfigure}
    \begin{subfigure}[t]{.49\linewidth}
        \centering
	\includegraphics[width=\linewidth, trim=20 0 0 0, clip]{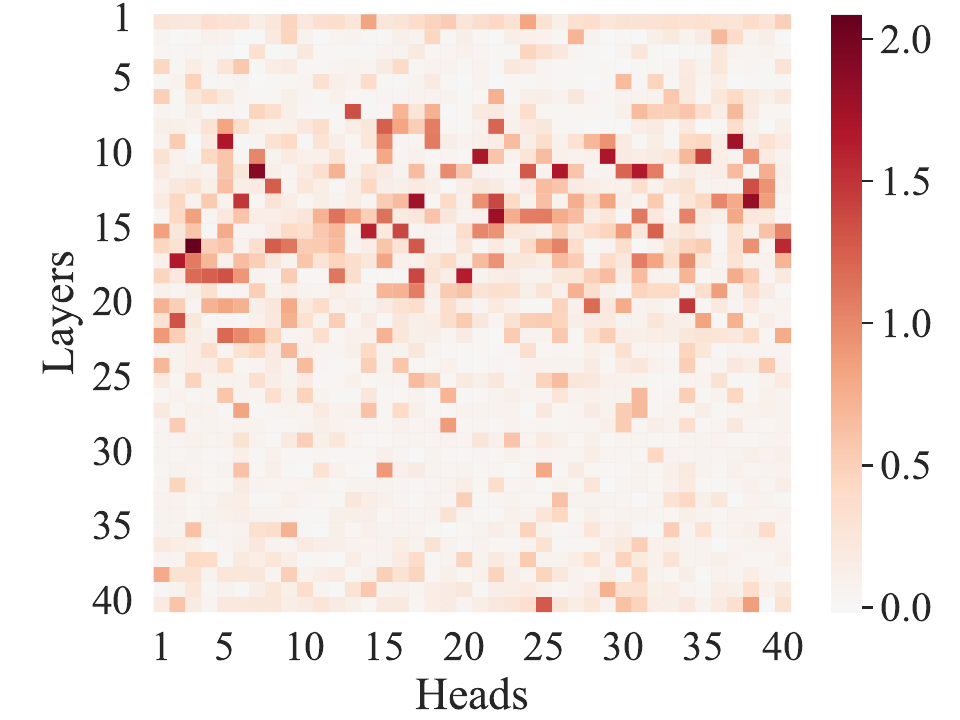}
        \caption{CSQA}
    \end{subfigure}
    \\
    \caption{Attention tracing results across different attention heads on Llama2-13B.}
    \label{fig:attn_comp}
\end{figure}

\begin{figure*}[t] 
    \centering
    \begin{subfigure}[t]{.245\linewidth}
        \centering
	\includegraphics[width=\linewidth]{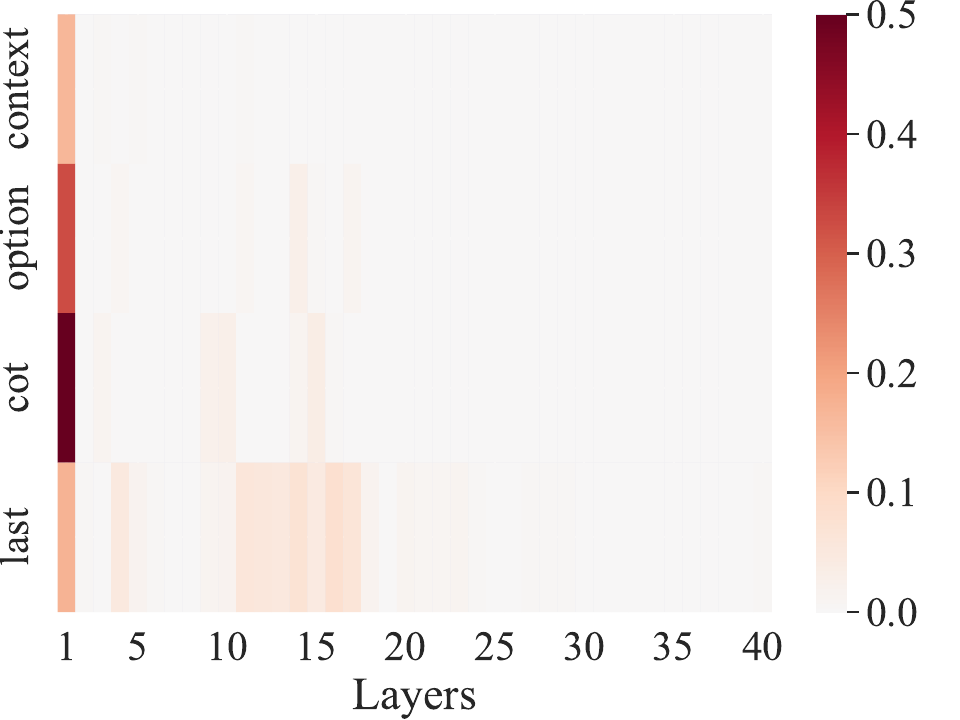}
        \caption{Correct Case's Attn}\label{fig:inter_wino_attention_cor}
    \end{subfigure}
    \begin{subfigure}[t]{.245\linewidth}
        \centering
	\includegraphics[width=\linewidth]{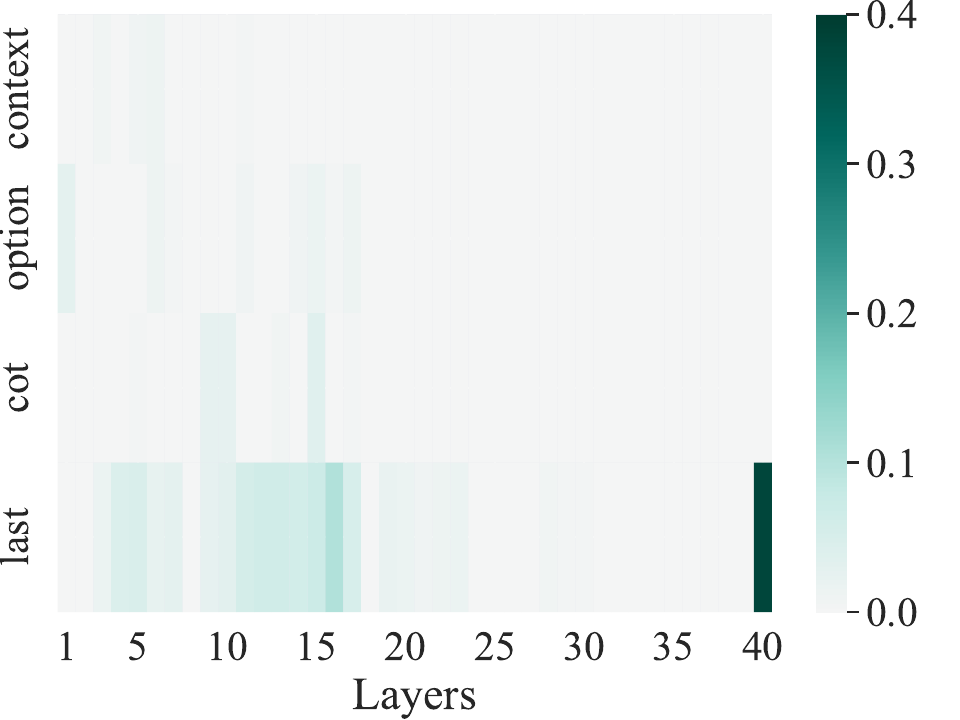}
        \caption{Correct Case's MLP}\label{fig:inter_wino_mlp_cor}
    \end{subfigure}
     \begin{subfigure}[t]{.245\linewidth}
        \centering
	\includegraphics[width=\linewidth]{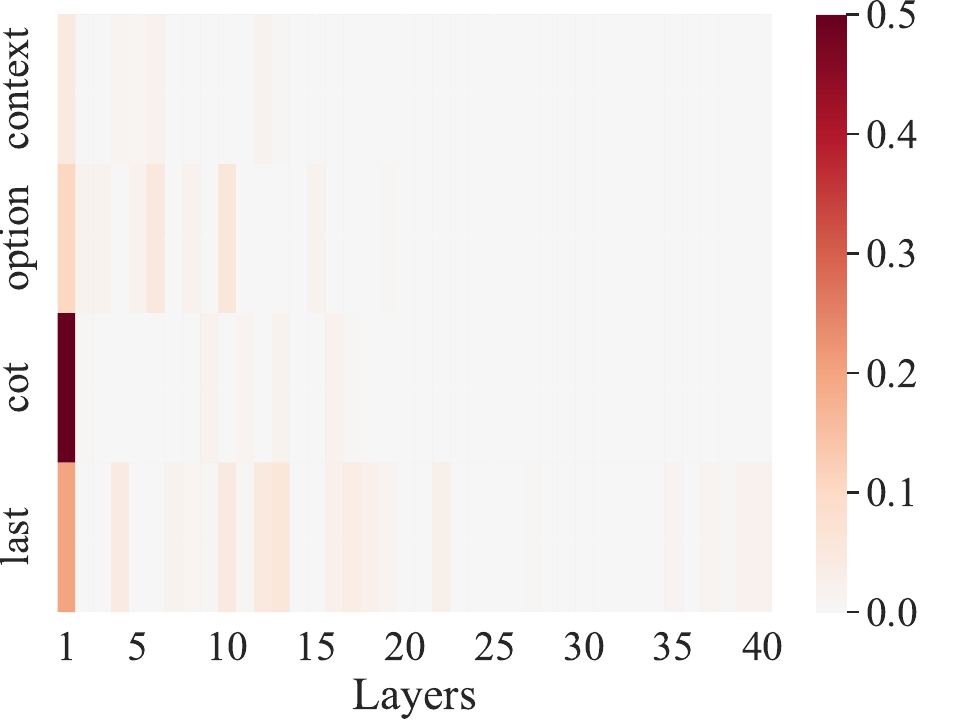}
        \caption{Drifting Case's Attn}\label{fig:inter_wino_attention_dri}
    \end{subfigure}
    \begin{subfigure}[t]{.245\linewidth}
        \centering
	\includegraphics[width=\linewidth]{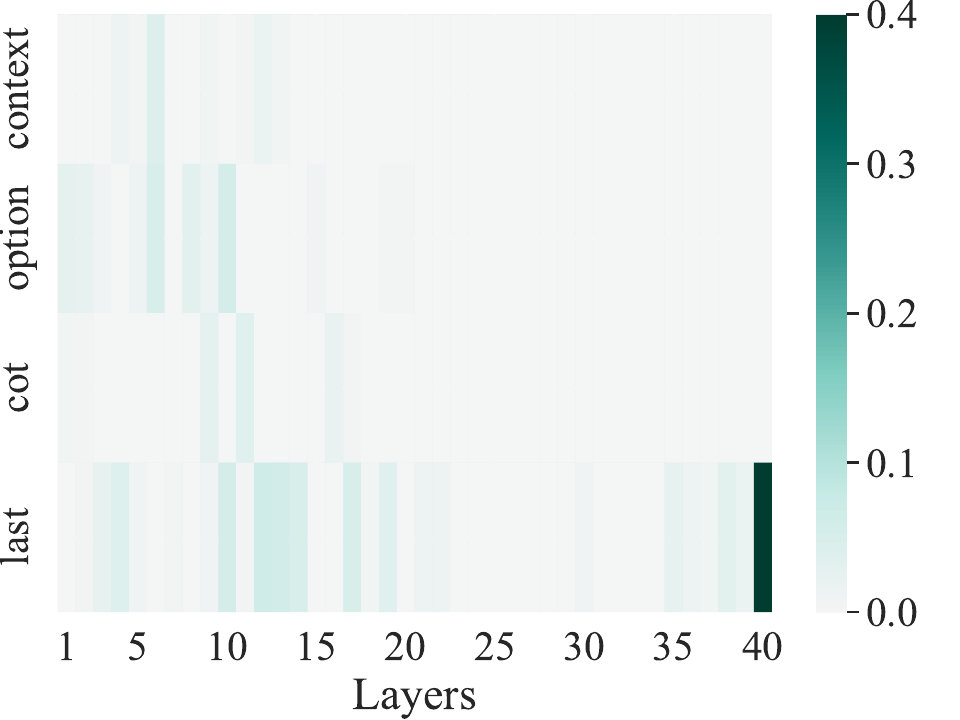}
        \caption{Drifting Case's MLP}\label{fig:inter_wino_mlp_dri}
    \end{subfigure}
    \\
    \caption{Intervention tracing results on Winogrande in correct and drifting answering cases.}
    \label{fig:wino_inter}
\end{figure*}

\subsection{Attention Tracing Experiment}\label{sec:3.3}
\paragraph{Experimental Settings}
We also design an experiment based on attention scores to validate Hypothesis \ref{hypo1} from another perspective. For a pair of rationales <$c, c^*$> targeting the same question context $q$ (c denotes the correct CoT and $c^*$ is the drifting one), we compute their attention divergence:
\begin{equation}\label{eq:attn_score}
\footnotesize{
    \begin{aligned}
    Attn(c | c^*) = \sum_{(i,j)\in C_{qc}}\frac{A^{(l,h)}_{i,j}}{|c|} - \sum_{(i,j)\in C_{qc^*}}\frac{A^{(l,h)}_{i,j}}{|c^*|}
    \end{aligned}}
\end{equation}
Here, we replace the $Attr(A_{i,j}^{(l)})$ in Equation \ref{eq:infor_flow} with the weights on the $h$-th attention matrix head $A_{i,j}^{(l,h)}$ and repeat the calculation in Equation $3$. 

\paragraph{Result and Analysis}
The results on two datasets are shown in Figure \ref{fig:attn_comp} and Figure \ref{fig:attn_comp_baichuan}. We can get the following observations: (1) The attention divergence is greater than 0 in most heads, which indicates \textbf{a lack of information from the question context in Rationale Drift cases (consistent with Claim 1).} (2) \textbf{The largest attention divergence appears around layer 15, which is consistent with the sites we find in Claim 2.} This once again illustrates that attention heads of these layers are crucial for the LLM to obtain contextual information when generating CoT.

\section{Tracing Information Flow in Answer}\label{sec:4}
In this section, our goal is to verify the information loss based on Hypothesis \ref{hypo2}. To achieve this, we first introduce the main tracing method in this section, which is called the causal tracing method ($\S$\ref{sec:4.1}). Next, by employing it, we trace the information flow in the answer generation stage and identify the mechanism behind the Answer Drift issue through comparative experiments ($\S$\ref{sec:4.2}). At last, we apply the attribution tracing method to verify our hypothesis from another perspective ($\S$\ref{sec:4.3}).

\subsection{Tracing Method}\label{sec:4.1}
 Since the task we study is in the form of multiple-choice questions, we set our focus on the feedforward pass that predicts the label. Inspired by the previous works \citep{rome,math,recal_fact}, we take the causal tracing method to quantify the contribution of different intermediate variables during this pass. Specifically, for hidden states $h_i^{(l)}$ in a clean run that predicts the answer, we have:
\begin{equation}
   \footnotesize{ \begin{aligned}
    &h_i^{(l)} = h_i^{(l-1)} + a_i^{(l)} + m_i^{(l)} \\
    &a_i^{(l)} = attn^{(l)}(h_1^{(l-1)}, ... , h_i^{(l-1)} ) \\ 
    &m_i^{(l)} = mlp^{(l)}(a_i^{(l)}+h_i^{(l-1)})
    \end{aligned}}
\end{equation}
where $i, l$ is the $i$-th token in the $l$-th layer, $a_i^{(l)}, m_i^{(l)}$ represents the activations of attention and MLP modules in Transformer \citep{transformer}. Supposing that a certain input part is represented as $z = [v_i^{(l)},  ..., v_j^{(l)}]$ after passing through a model component, we set $v_k^{(l)} = v_k^{(l)} + \epsilon$ for $k \in [i, j]$ to intervene this hidden vector, where $\epsilon$ is Gaussian noise.\footnote{We select $\epsilon$ to be 3 times larger than the empirical standard deviation of hidden embeddings in each dataset.} Thus, we can compute the direct effect (DE) of this component:
\begin{equation}
   \footnotesize{ \begin{aligned}
    DE(z) = \frac{P(o) - P^*_z(o)}{P(o)}
    \end{aligned}}
\end{equation}
where $P(o)$ is the probability of the model's final prediction, $P^*_z(o)$ is the probability after the intervention. Therefore, through this metric, we can quantify the contributions of different components in changing the final prediction, thereby tracing the information flow in this stage.

\subsection{Intervention Tracing Experiment}\label{sec:4.2}
\paragraph{Experimental Settings}
We sample correct and drifting answering cases from datasets, average over them and compute the average direct effect (ADE). Here we compute the impact of four components on the final prediction: context (question contexts), option (question options), CoT, and last (the last token before the label prediction).
\paragraph{Results and Analysis}
We report the result of Llama2-13B on Winogrande in Figure \ref{fig:wino_inter} and others in Appendix \ref{sec:inter_trac}, from which we can get two conclusions: \textbf{(Claim 3) For attention modules, drifting cases loss information from the question.} When the answer is correct, we can observe a high effect on the context and option in the first layer (see Figure \ref{fig:inter_wino_attention_cor}). But for the drifting case, the LLM extracts limited information at these positions (see Figure \ref{fig:inter_wino_attention_dri}). This aligns with Hypothesis \ref{hypo2}. \textbf{(Claim 4) For MLP modules, the information is not lost.} We observe the same high-effect sites in the last layer and shallow layers of the last token, they do not show regular differences (see Figure \ref{fig:inter_wino_mlp_cor} and \ref{fig:inter_wino_mlp_dri}).

\begin{figure}[tbp] 
    \centering
    \begin{subfigure}[t]{.49\linewidth}
        \centering
	\includegraphics[width=\linewidth]{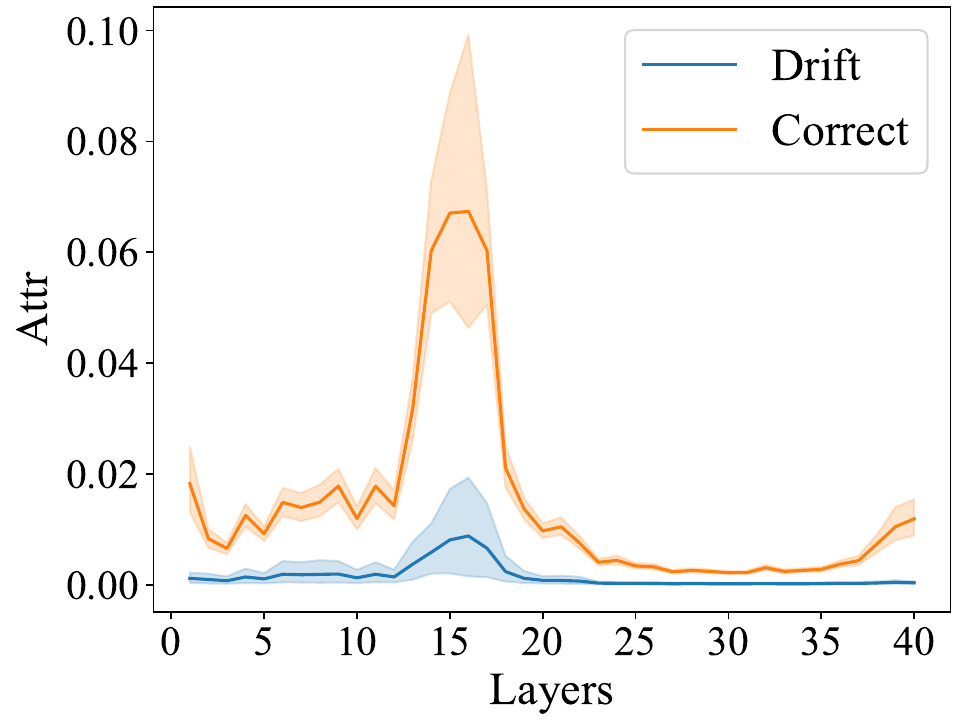}
        \caption{Winogrande}
    \end{subfigure}
    \begin{subfigure}[t]{.49\linewidth}
        \centering
	\includegraphics[width=\linewidth]{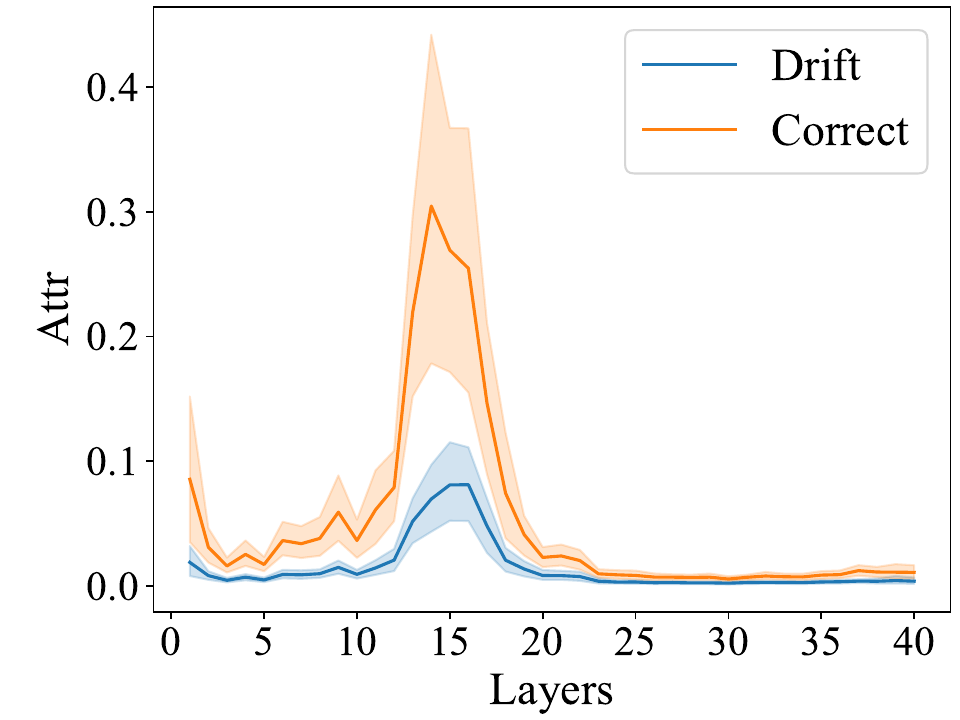}
        \caption{CSQA}
    \end{subfigure}
    \\
    \caption{Attribution tracing results on Llama2-13B during the answer generation stage.}
    \label{fig:attr_label}
\end{figure}

\subsection{Attribution Tracing Experiment}\label{sec:4.3}
\paragraph{Experimental Settings}
For further validation of our hypothesis, we also use the attribution score in $\S$\ref{sec:3} to trace the information flow in this stage. Referring to Equation \ref{eq:infor_flow}, we compute the score between the question context and the last token (since it's used for generating the answer). We set the $F(\cdot)$ in Equation \ref{eq:attr} as the loss for predicting the final answer, comparing the scores for correct and drifting cases after averaging across samples. 

\paragraph{Results and Analysis}
The results of this experiment are reported in Figure \ref{fig:attr_label} and \ref{fig:attr_label_b}. We can observe that, \textbf{when the Answer Drift issue occurs, the information flow from the question significantly decreases}. This verifies the information loss we mention in Hypothesis \ref{hypo2} and Claim 3.

\begin{figure*}[tbp]
    \centering
    \includegraphics[width=\textwidth]{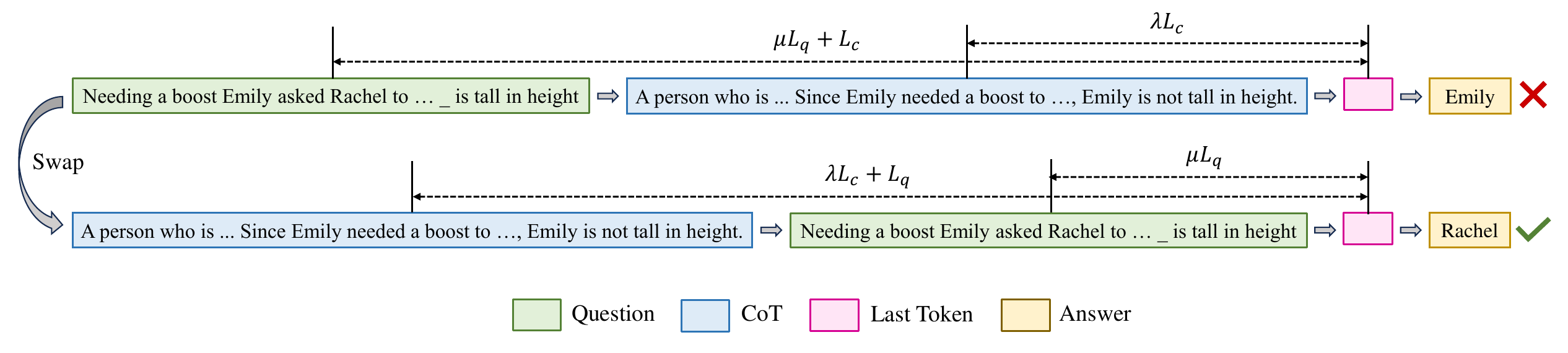}
    \caption{An example of our serial-position swap method.}
    \label{fig:sps}
\end{figure*}

\section{Mitigating Toxic CoT Problem}\label{sec:5}
In this section, we propose a novel method called $\mathbb{RIDERS}$ (\textbf{R}esidual decod\textbf{I}ng and s\textbf{ER}ial-position \textbf{S}wap) to address the Toxic CoT problem. We first introduce the two components in it, which are designed based on Hypothesis \ref{hypo1} and \ref{hypo2}, respectively ($\S$\ref{sec:5.1}). Then, we conduct experiments on commonsense reasoning benchmarks, demonstrating the effectiveness of our method ($\S$\ref{sec:5.2} and $\S$\ref{sec:5.3}). At last, we conduct extra experiments to further emphasize the contribution of our approach ($\S$\ref{sec:5.4}).

\subsection{Mitigation Method}\label{sec:5.1}
\paragraph{Residual Decoding} We design a new decoding methodology to address the Rationale Drift issue, in which we construct a virtual residual structure during the CoT generation, ``connecting'' the question context with each CoT token. Our decoding algorithm is demonstrated in Algorithm \ref{alg:RD}. In each iteration of generating a new token, we first select the top $n$ tokens with the highest probabilities and record their logits scores (line 3). Then we calculate the attention score between the context and current token like Equation \ref{eq:attn_score}, normalize it, and add it as an additional reward to promote more information flow (lines 6,7). Finally, we select the token with the highest score to update the input and repeat the process until the termination condition is met. We use the attention matrix in layer 15 to compute the attention score, since it is crucial for the exchange of contextual information according to Claim 2. More implementation details of this method are provided in Appendix \ref{sec:main_method}.
\begin{algorithm}[tbp]
	\caption{Residual Decoding Algorithm}
	\label{alg:RD}
	\footnotesize{\begin{algorithmic}[1]
		\REQUIRE model $\mathcal{M}$, input $x$, question context $q$, candidate\_num $n$, weight $\omega$.
        \FOR{iteration $i \in 0, 1, ...$}
            \STATE $logits = \mathcal{M}(x)$
            \STATE $tokens$, $scores$ = top($logits, n$)
            \FOR{$j \in 1, ... n$}
                \STATE $t = tokens[i]$
                \STATE $attn\_score = $ Attn($q$, $t$) / Attn($x$, $t$)
                \STATE $scores[j] = scores[j]$ + $\omega * attn\_score$
            \ENDFOR
            \STATE $idx$ = argmax($scores$)
            \STATE $t = tokens[idx]$
            \STATE $x = x + t$
            \IF{stop($t$)}
            \STATE \textbf{break}
            \ENDIF
        \ENDFOR
		\STATE \textbf{return} $x$
	\end{algorithmic} } 
\end{algorithm}

\paragraph{Serial-Position Swap}
In this method, we attempt to compensate for the information lack in the Answer Drift issue. According to previous research on the serial-position effect in context, models tend to utilize information better at the beginning and end of the input \citep{long_range,lost_in_middle}. In our work, the beginning of the input are prompts, while the current question and the generated CoT are both located at the end. Therefore, when they are closer to the last token, their information is more easily utilized in the final prediction. As in Figure \ref{fig:sps}, we denote the lengths of the question and CoT as $L_q$ and $L_c$, and assume that the key information is located at positions $\mu L_q$ and $\lambda L_c$  (similar to the center of mass in physics). We can infer that, after the swapping operation in Figure \ref{fig:sps}, the distance from the question to the end is reduced ($\mu L_q + L_c \rightarrow \mu L_q$). Besides, if we consider the total distance from the question and CoT to the end, we can perform the following calculation:
\vspace{-0.15mm}
\begin{equation}
  \footnotesize{  \begin{aligned}
     &d_1 = \mu L_q + L_c + \lambda L_c \\
     &d_2 = \lambda L_c + L_q + \mu L_q \\
     &\Delta d  = d_2 -d_1 = L_q - L_c
    \end{aligned} }
\end{equation}
where $d_1$ is the total distance in normal serial positions and $d_2$ is the distance after swapping the two components. In most scenarios, we have $L_q < L_c$, thus, we can infer that $\Delta d < 0$. That means, if we replace the original order of ``\textit{[Question] + [CoT]}'' with the order of ``\textit{[CoT] + [Question]}'', we can not only increase the intensify of information flow from the question to the final prediction, but also reduce the total information loss due to the reduction in total distance. Although this method is straightforward in implementation, it proves to be effective in both theory and experiments.

\begin{table*}[t]
\centering
 \scalebox{0.8}{
\begin{tabular}{llcccccccccccc}
\toprule
\multicolumn{2}{l}{\multirow{2}{*}{\textbf{Method}}} & \multicolumn{2}{c}{\textbf{Winogrande}} & \multicolumn{2}{c}{\textbf{CSQA}} & \multicolumn{2}{c}{\textbf{HellaSwag}} & \multicolumn{2}{c}{\textbf{SIQA}} & \multicolumn{2}{c}{\textbf{PIQA}} & \multicolumn{2}{c}{\textbf{Avg}} \\
\cline{3-14}
& & ACC$\uparrow$ & TR$\downarrow$ & ACC$\uparrow$ & TR$\downarrow$& ACC$\uparrow$ & TR$\downarrow$ & ACC$\uparrow$ & TR$\downarrow$& ACC$\uparrow$ & TR$\downarrow$ & ACC$\uparrow$ & TR$\downarrow$  \\
\midrule
\multicolumn{2}{l}{Few-shot Answer} & 57.1  & - & 66.8 & - & 45.3& - & 68.2 & - & 61.6 & - & 59.8 & -\\
\multicolumn{2}{l}{Chain-of-Thought} & 56.7  & 43.1 & 69.9 & 28.8 & 45.4 & 34.8 & 67.3 & 37.3 & 62.7 & 41.3 & 60.4 & 37.1\\
\multicolumn{2}{l}{Self-Consistency} & 56.4  & 40.9 & 72.4 & 24.9 & 46.2& 33.7 &64.7 & 41.0 & 54.6 & 45.8 & 58.9 & 37.3\\ 
\multicolumn{2}{l}{Self-Refine}  &48.7 & 50.3 & 55.5 & 47.0 & 42.0 & 37.2 & 65.1 & 38.8 & 50.8 & 48.3 & 52.4 & 44.3\\
\multicolumn{2}{l}{Least-to-Most} & 58.9 & 38.8 & 69.0 & 27.5 & 31.3 & 39.4 & 66.3 & 36.6 & 65.7 & 32.2 & 58.2 & 34.9\\
\multicolumn{2}{l}{Contrasive CoT} & 59.4 & 38.0 & 71.1 & 23.2 & 45.9 & 35.6 & 67.0 & 36.0 & 68.1 & 44.9 & 62.3 & 35.5\\ 
\midrule
\multirow{3}{*}{\textbf{Ours}} &RD Only & 58.6 & 29.5 & 72.2 & 13.5& 49.4 & 24.0 & 69.4 & 20.8 & 66.3 & 26.2 & 63.0 & 22.8\\
&SPS Only &59.1 & 22.8 & 72.6 & 18.9 & 49.3 & 18.7 & 69.7 & 28.8 & 68.5 & 27.6 & 63.7 & 23.4\\
&RIDERS & \textbf{60.7} & \textbf{16.6}& \textbf{73.2} & \textbf{6.7} & \textbf{50.6} & \textbf{12.8} &  \textbf{71.9} & \textbf{15.3} & \textbf{69.9} & \textbf{15.9} & \textbf{65.3} & \textbf{13.5}\\ 
\bottomrule
\end{tabular}}
\caption{Performance comparison across five commonsense reasoning datasets on Llama2-13B.}
\label{tab:main}
\end{table*}

\subsection{Mitigation Experimental Settings}\label{sec:5.2}
\paragraph{Datasets} Following previous works, we use five representative commonsense reasoning benchmarks: \textbf{WinoGrande} \citep{winogrande}, \textbf{CSQA} \citep{csqa}, \textbf{HellaSwag} \citep{hella}, \textbf{SIQA} \citep{siqa} and \textbf{PIQA} \citep{piqa}. The specific information of each dataset is reported in Appendix \ref{sec:main_exp}. 
\paragraph{Metrics} In addition to the commonly used \textbf{Accuracy (ACC)} metric, we also introduce a new metric --- \textbf{Toxic Rate (TR)}, to quantify the severity of Toxic CoT problems:
\begin{equation}\label{eq:tr}
    \footnotesize{   \begin{aligned}
    TR(f) = |C_d \cap W_f| / |W_f|
    \end{aligned}}
\end{equation}
 where $C_d$ denotes questions that models give correct answers directly and $W_f$ denotes questions that models give wrong answers after applying method $f$. Thus, we can infer that the lower the TR, the fewer Toxic CoT problems the method introduces.

\paragraph{Baselines} As our research focuses on enhancing CoT methods in commonsense reasoning, we select some of the latest CoT-like methods applicable to this task for comparison: \textbf{Few-shot Answer}, \textbf{Chain-of-Thought} \citep{cot}, \textbf{Self-Consistency} \citep{sc}, \textbf{Self-Refine} \citep{sr}, \textbf{Least-to-Most} \citep{l2m} and \textbf{Contrasive CoT} \citep{contrast_cot}. For all methods, we employ a 5-shot prompt and use 4 NVIDIA GeForce RTX 3090 GPUs for inference. More implementation details can be found in the Appendix \ref{sec:main_exp}.

\subsection{Mitigation Results}\label{sec:5.3}
The main result of our experiments on Llama2-13B is shown in Table \ref{tab:main}. (The results on more models are presented in Appendix \ref{sec:main_exp}.) We can get the following conclusions: \textbf{(1) Our method effectively mitigates Toxic CoT problems.} 
Compared to CoT prompting, our method reduces the Toxic Rate by an average of \textbf{23.6\% } across five datasets. Besides, compared to other advanced CoT-like methods, our method causes the fewest Toxic CoT problems (decreased by an average of \textbf{22.0\%} over SOTA methods). \textbf{(2) Our method can also improve the model's overall performance on commonsense reasoning.} Our work improves the accuracy on all benchmarks (improved by \textbf{5.5\%} compared to CoT and \textbf{3.0\%} compared to SOTA methods on average). This proves that the Toxic CoT problem poses a bottleneck in LLM's commonsense reasoning, highlighting the value of our work.

\begin{table}[tbp]
\centering
 \scalebox{0.8}{
\begin{tabular}{lcccc}
\toprule
\multirow{2}{*}{\textbf{Method}} & \multicolumn{2}{c}{\textbf{Winogrande}} & \multicolumn{2}{c}{\textbf{CSQA}}\\
\cline{2-5}
 & Type1 & Type2 & Type1 & Type2  \\
\midrule
CoT & 0.0  & 0.0 & 0.0 & 0.0  \\
RD Only& 41.9 & 47.8 & 56.7 & 46.2 \\
SPS Only &58.1 & 78.3 & 46.7 & \textbf{92.3}  \\
RIDERS & \textbf{74.2} & \textbf{82.6}& \textbf{73.3} & 84.6\\ 
\bottomrule
\end{tabular}}
\caption{Accuracy on the two types of drifting issues.}
\label{tab:drift}
\end{table}

\subsection{Discussion and Analysis}\label{sec:5.4}
\paragraph{Performance on Two Drifting Issues}
To demonstrate the effectiveness of our method in addressing the Rationale Drift issue (Type1) and Answer Drift issue (Type2), we conduct experiments on these samples and report the results in Table \ref{tab:drift}. Both of our methods can mitigate the corresponding issues (RD solves \textbf{49.3\%} Rationale Drift issue on average, while SPS resolves \textbf{85.3}\% Answer Drift issue on average). This verifies the validity of our hypothesis \ref{hypo1}, \ref{hypo2}, as all of these methods are built upon them. Besides, combining the two methods leads to even greater improvements, demonstrating the necessity of optimizing from both of these perspectives.
\begin{figure}[tbp] 
    \centering
    \begin{subfigure}[t]{.49\linewidth}
        \centering
	\includegraphics[width=\linewidth]{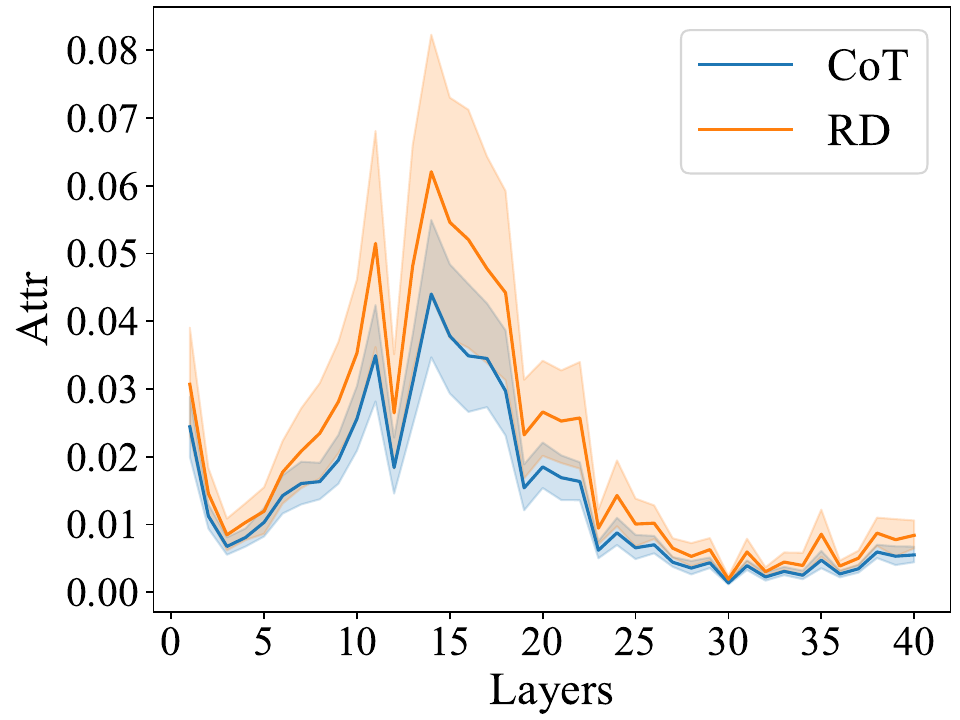}
        \caption{RD method}
    \end{subfigure}
    \begin{subfigure}[t]{.49\linewidth}
        \centering
	\includegraphics[width=\linewidth]{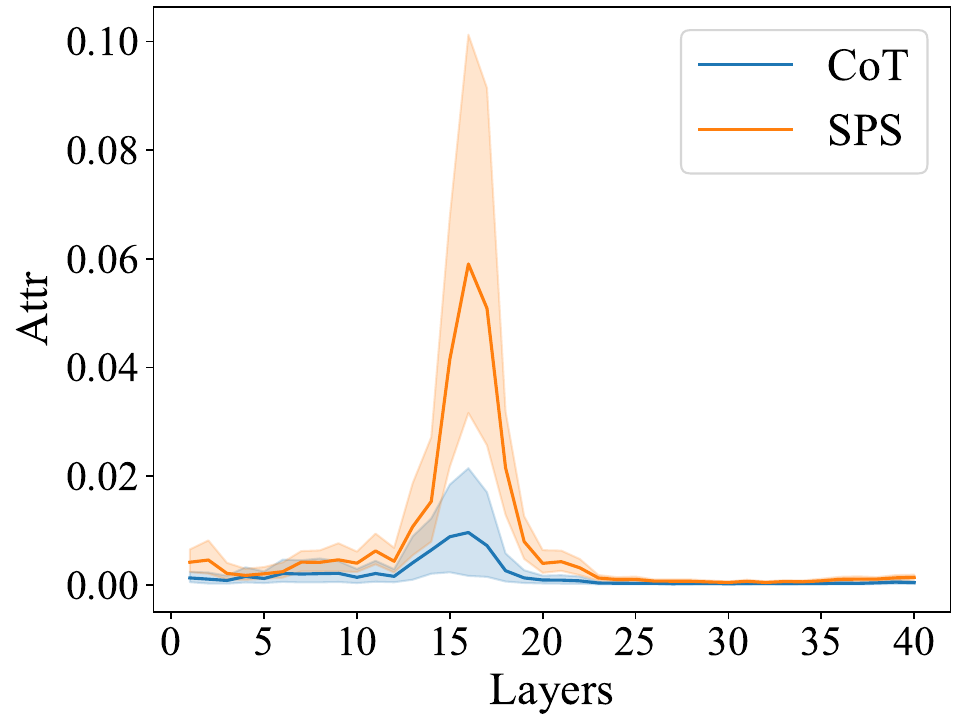}
        \caption{SPS method}
    \end{subfigure}
    \\
    \caption{Information flow comparison on Winogrande after applying our two methods.}
    \label{fig:attr_method_wino}
\end{figure}

\paragraph{Performance in the Model}
In $\S$\ref{sec:3} and $\S$\ref{sec:4}, we probe information loss in two issues by tracing the information flow in models. Here, we repeat the attribution tracing experiments, comparing the differences before and after applying our method to further validate the effectiveness of our work. As we can see from Figure \ref{fig:attr_method_wino} and \ref{fig:attr_method_csqa}, our two methods (orange lines) increase the information flow from questions in two stages compared to CoT prompting (blue lines). This indicates our method indeed compensates for the information loss in the LLM.

\begin{table}[tbp]
\centering
 \scalebox{0.8}{
\begin{tabular}{lcccc}
\toprule
\multirow{2}{*}{\textbf{Method}} & \multicolumn{2}{c}{\textbf{ProofWriter}} & \multicolumn{2}{c}{\textbf{FOLIO}}\\
\cline{2-5}
& ACC$\uparrow$ & TR$\downarrow$ & ACC$\uparrow$ & TR$\downarrow$  \\
\midrule
CoT & 	56.8  & 19.9 & 45.1 & 11.6  \\
RIDERS & \textbf{78.2} & \textbf{6.4}& \textbf{48.0} & \textbf{7.5}\\ 
\bottomrule
\end{tabular}}
\caption{Performance comparison across two logical reasoning tasks on Llama2-13B.}
\label{tab:other_task}
\end{table}

\paragraph{Performance on Other Tasks}
Our work mainly focuses on commonsense reasoning tasks. To further evaluate the effectiveness of our approach on other types and forms of tasks, we conduct the main experiments on two logical reasoning tasks: \textbf{ProofWriter} \citep{proofwriter} and \textbf{FOLIO} \citep{folio}). As presented in Table \ref{tab:other_task}, we can find that the Toxic CoT problem also exists in other tasks, and our method can still mitigate it.

\begin{table}[tbp]
\centering
 \scalebox{0.7}{
\begin{tabular}{lcccccc}
\toprule
\textbf{Method} & \textbf{Wino} & \textbf{CSQA} & \textbf{Hella} & \textbf{SIQA} & \textbf{PIQA}  & \textbf{Avg} \\
\midrule
CoT & 3.2k & 2.6k & 4.2k & 2.9k  & 2.7k & 3.1k \\ 
SC & 4.5k & 3.2k & 5.5k & 3.8k & 4.3k & 4.3k \\
SR & 9.2k & 6.9k & 10.2k & 7.6k & 6.5k & 8.1k \\
L2M & 8.7k & 7.9k & 11.4k & 8.3k & 6.7k & 8.6k \\
CON & 4.6k & 3.5k & 5.4k & 3.8k & 3.5k & 4.2k \\ 
\midrule
Ours & 3.6k & 2.9k & 4.9k & 3.4k & 3.3k & 3.6k \\
\bottomrule
\end{tabular}}
\caption{Token consumption per example comparison.}
\label{tab:cost}
\end{table}

\paragraph{Cost Analysis}
For the applicability, we measure the computation and time cost of our approach. Here we compare the token cost between our method and the baseline. According to Table \ref{tab:cost}, our method requires fewer tokens compared to other SOTA methods (only $1.2\times$ cost of the basic CoT method). We also compare the time cost of our decoding method in Appendix \ref{sec:cost} and find that the speed of RD is comparable to existing decoding strategies. Therefore, we illustrate the cost-efficiency of our approach across different tasks.

\section{Related Work}
\subsection{CoT Problems Analysis and Mitigation}
Recently, many works have focused on analyzing and mitigating problems in CoT reasoning. For analytical work, most studies focus on black-box LLMs. Through intervening or paraphrasing prompts and comparing outputs, researchers can interpret the reasons leading to errors in the model's reasoning \citep{measure-faithful,score,cot-matters}. For optimization works, they design additional supervision signals or training processes for the model  \citep{tailor,crystal} or leverage external resources for the model \citep{reflexion,rethink_retrival,faithful-cot}. However, these works lack the probing of inner mechanisms behind these problems, leading to insufficient analysis or less universally applicable optimization methods.

\subsection{Mechanistic Interpretability}
The work on mechanistic interpretability aims to understand the internal mechanisms of models when performing various tasks. Early work focused on how the model stores factual knowledge internally \citep{rome,knowledgeneuron}. In recent times, some research efforts have shifted towards examining how models retrieve and utilize knowledge. This includes internal knowledge retrieval \citep{recal_fact}, knowledge retrieval from prompts \citep{labelworlds,cuthead_zhuoran}, and the utilization of knowledge for reasoning purposes, such as math reasoning \citep{math} and multi-step reasoning \citep{multi-step_mechan}. However, there is limited existing work that explains commonsense reasoning and CoT reasoning, which are significant contributions of this work.

\section{Conclusion}
In this paper, we find a problem named Toxic CoT, which results in the model's reasoning deviating from the original correct answer when utilizing CoT-like prompting. Through tracing the internal information flow of the LLM with attribution tracing and causal tracing methods, we prove that this problem is mainly caused by the model's lack of information from the question in shallow attention layers when generating rationales or answers. Based on this result, we propose the $\mathbb{RIDERS}$ method to mitigate the Toxic CoT problem from both decoding and serial position perspectives. Through extensive experiments on multiple commonsense reasoning datasets, we verify the effectiveness of our approach in mitigating Toxic CoT problems and enhancing the model's overall commonsense reasoning capabilities. 

\section*{Limitations}
Although our work conducts an in-depth interpretation and mitigation of the Toxic CoT problem, it has several limitations. Firstly, like former commonsense reasoning works \citep{rainer,crystal,self_eval}, our research focuses on the form of multi-choice questions. This stems from the absence of effective evaluation methods for open-ended commonsense reasoning, leading to the predominance of benchmarks in this format. This calls for advancements in benchmark-related research. Secondly, we refrain from analyzing Toxic CoT problems in more reasoning tasks such as math, primarily due to the poor performance of current moderately-sized white-box models on these tasks. For instance, Llama2-13B achieves a mere 7.2\% accuracy on GSM8K \citep{gsm8k} without utilizing the CoT technique. This calls for developments in model-related research. We leave these limitations as our future work to explore.

\section*{Acknowledgement}
This work is supported by the Strategic Priority Research Program of Chinese Academy of Sciences (No. XDA27020203), the National Natural Science Foundation of China (No. 62176257,62276095).

\bibliography{custom}

\clearpage

\appendix

\section{Early Statistical Experiments}
\label{sec:early-stat}
In this section, we conduct early experiments on existing representative commonsense reasoning datasets to analyze the prevalence of Toxic CoT problems through statistical methods.
\paragraph{Datasets}
We utilize five representative common-sense reasoning datasets to analyze the distribution of Toxic CoT problems. The basic information of the dataset is outlined in Table \ref{tab:data_stat}. It is noteworthy that, owing to the extensive size of Hellaswag's dev set (over 10,000), we extract 2,000 instances for the experiment.
\paragraph{Metric}
We design a new metric called Toxic Rate, which measures the proportion of Toxic CoT problems among all errors. Its calculation method is shown in Equation \ref{eq:tr}.
\paragraph{Results}
The result of our early statistical experiments is reported in Table \ref{tab:Toxic_stat}. Here we use \textbf{\textit{Llama2-13B-Chat-hf}} to present the white-box LLM and use \textit{\textbf{GPT-3.5-turbo-1106}} to present the black-box-model. The average Toxic Rates are as high as \textbf{37.0\%} and \textbf{32.8\%} across the five datasets, indicating that this issue cannot be ignored and warrants further investigation.

\begin{table*}[b]
\centering
\begin{tabular}{lccccc}
\toprule
 & \textbf{Winogrande} & \textbf{CSQA} & \textbf{HellaSwag} & \textbf{SIQA} & \textbf{PIQA}\\
\midrule
Split & dev  & dev & dev & dev & dev  \\
\#Sample Num & 1267 & 1221 & 2000 & 1954 & 1838 \\
\#Option Num & 2 & 5 & 4 & 3 & 2 \\
\bottomrule
\end{tabular}
\caption{Dataset information in this work.}
\label{tab:data_stat}
\end{table*}

\begin{table*}[b]
\centering
\begin{tabular}{lcccccc}
\toprule
 & \textbf{Winogrande} & \textbf{CSQA} & \textbf{HellaSwag} & \textbf{SIQA} & \textbf{PIQA} & \textbf{Avg}\\
\midrule
Llama2-13B & 43.1 & 28.8 & 34.8 & 37.3 & 41.2 & 37.0  \\
GPT-3.5 & 34.8 & 37.8 & 27.5 & 37.5 & 26.4 & 32.8 \\
\bottomrule
\end{tabular}
\caption{Toxic rate on different datasets and models.}
\label{tab:Toxic_stat}
\end{table*}


\section{Toxic Reason Statistical Experiments}
\label{sec:Toxic_reason}

In this section, we manually categorize the error types of Toxic CoT problems through statistical classification. Specifically, we sample 1,000 examples from CSQA and 1,000 examples from Winogrande, classifying the Toxic CoT problems (In all of the probing experiments in the main text, we use these samples as our probing data). The results are presented in Table \ref{tab:Toxic_reason}. In the inconsistent error, the model exhibits logical inconsistency with the preceding context when generating CoT or the final answers. In the factual error, the CoT contains incorrect factual knowledge, which leads to erroneous answers. The presence of question errors reflects the subpar quality of the dataset. In such cases, questions may exhibit multiple viable answers or all options are incorrect. As for the other error, the questions trigger certain refusal-to-answer mechanisms in the model (e.g., inquiries about how to commit murder), leading to the identification of incorrect answers.

As the inconsistency error constitutes the predominant portion of all reasons, our work focuses on addressing this issue. We further categorize this error into Rationale Drift and Answer Drift based on the error occurrence (see $\S$\ref{sec:2.2} for their definitions). 

\begin{table}[h]
\centering
\scalebox{0.95}{
\begin{tabular}{lcccccc}
\toprule
 \textbf{Reason} & \textbf{Winogrande} & \textbf{CSQA} \\
\midrule
Inconsistent Error & \textbf{71(78.9\%)}& \textbf{48(53.9\%)}\\
Factual Error & 16(17.8\%) & 17(19.1\%) \\
Question Error & 2(2.2\%) & 21(23.6\%) \\
Other Error & 1(1.1\%) & 3(3.4\%) \\
Sum & 90 & 89 \\
\bottomrule
\end{tabular}}
\caption{The classification of CoT reasoning errors}
\label{tab:Toxic_reason}
\end{table}

\section{More Details for Reasoning Tracing}
\label{sec:attr_trac}
\paragraph{Method Implementation} We introduce the attribution score method in $\S$ \ref{sec:3.1}. In Equation \ref{eq:attr}, we set $m$ = 20 following the previous works. For $F(\cdot)$, we set it as the language modeling loss (for next-token prediction) during the CoT generation. Here, we obtain this value directly from the output of the \textit{LlamaForCausalLM} module using the \textit{Transformers} library. In Equation \ref{eq:infor_flow}, we partition the step numbers $|N|$ in CoT based on the occurrence of periods in the text. For models, we use \texttt{Llama2-13B-Chat} and \texttt{Baichuan2-13B-Chat}.
\paragraph{Attribution Tracing Experiment}

\begin{figure*}[htbp] 
    \centering
    \begin{subfigure}[t]{.49\linewidth}
        \centering
	\includegraphics[width=\linewidth]{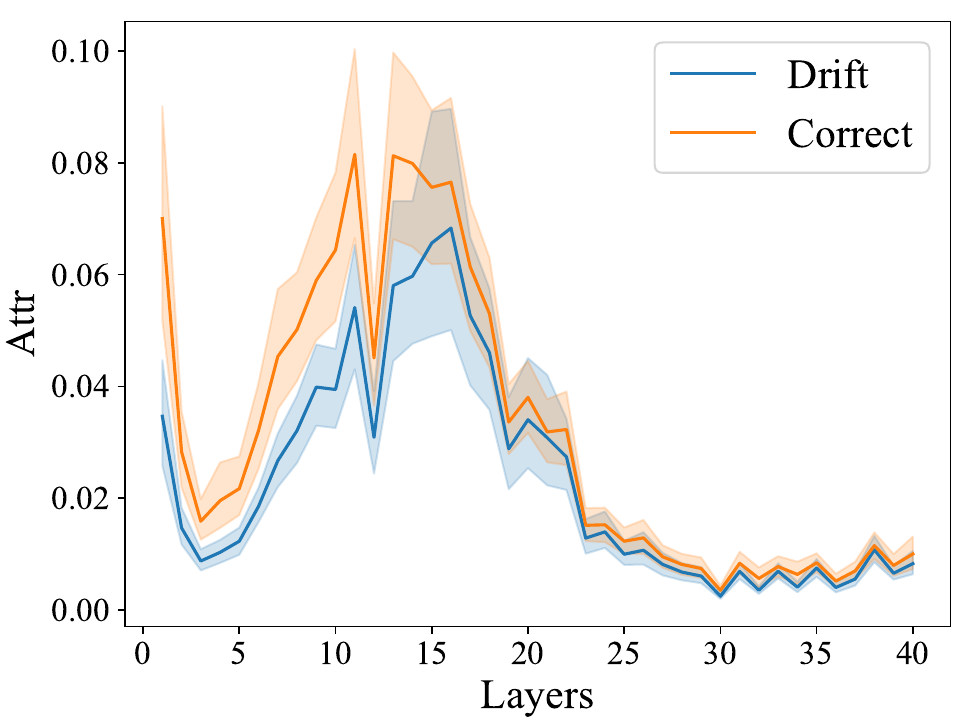}
        \caption{Llama2-13B}
    \end{subfigure}
    \begin{subfigure}[t]{.49\linewidth}
        \centering
	\includegraphics[width=\linewidth]{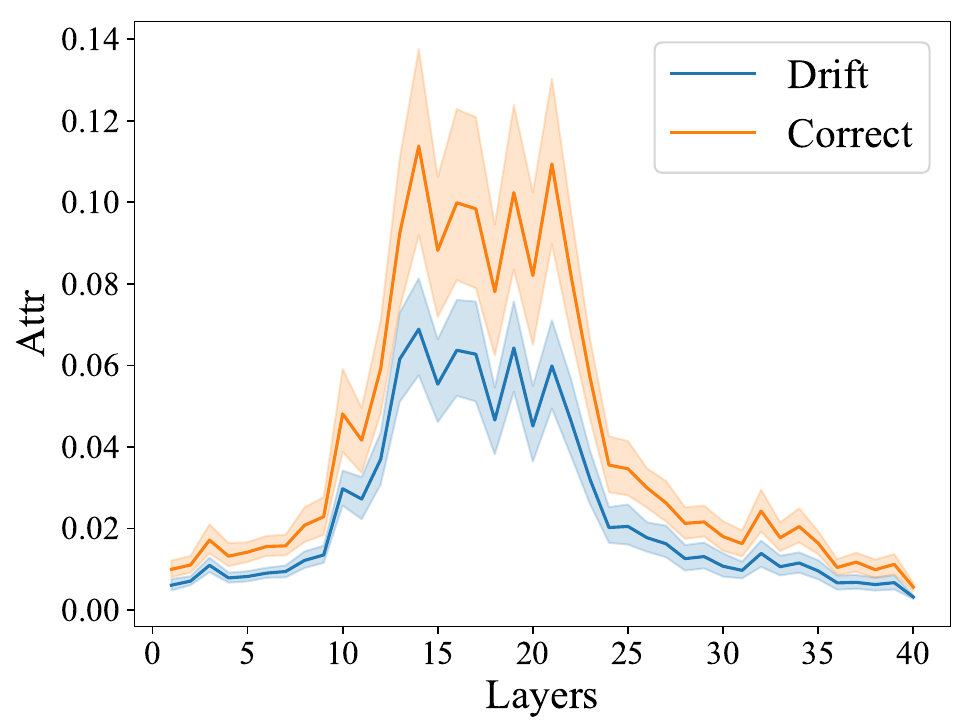}
        \caption{Baichuan2-13B}
    \end{subfigure}
    \\
    \caption{Attribution tracing results on CSQA.}
    \label{fig:attr_comp_csqa}
\end{figure*}

\begin{figure*}[htbp] 
    \centering
    \begin{subfigure}[t]{.49\linewidth}
        \centering
	\includegraphics[width=\linewidth]{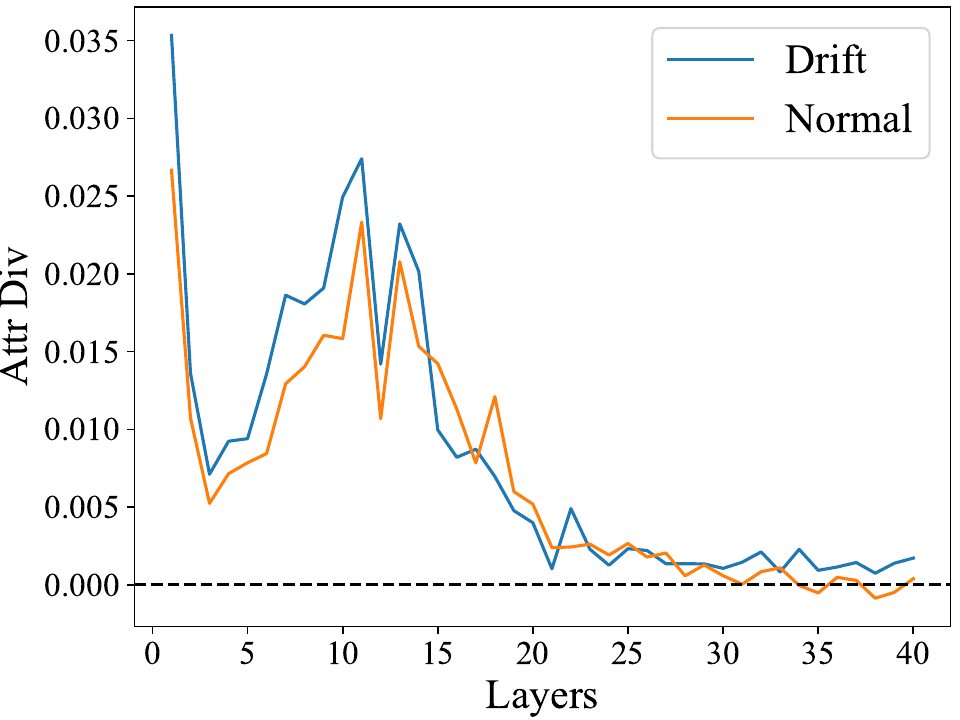}
        \caption{Llama2-13B}
    \end{subfigure}
    \begin{subfigure}[t]{.49\linewidth}
        \centering
	\includegraphics[width=\linewidth]{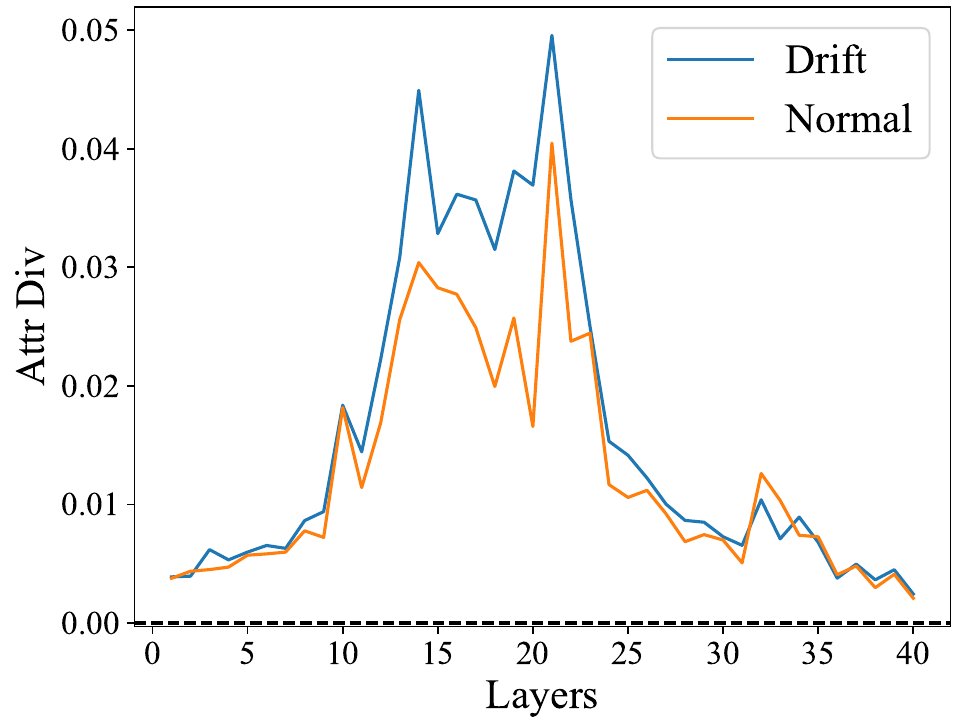}
        \caption{Baichuan2-13B}
    \end{subfigure}
    \\
    \caption{Information flow divergence comparison on CSQA.}
    \label{fig:attr_comp_csqa_extra_b}
\end{figure*}

\begin{figure*}[htbp] 
    \centering
    \begin{subfigure}[t]{.49\linewidth}
        \centering
	\includegraphics[width=\linewidth]{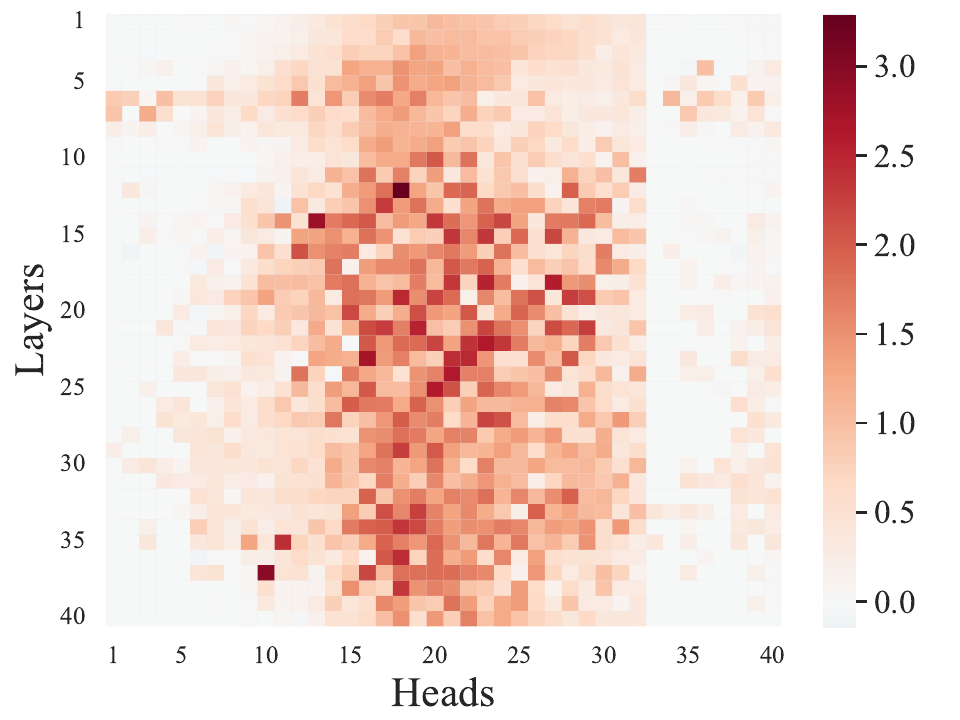}
        \caption{Winogrande}
    \end{subfigure}
    \begin{subfigure}[t]{.49\linewidth}
        \centering
	\includegraphics[width=\linewidth]{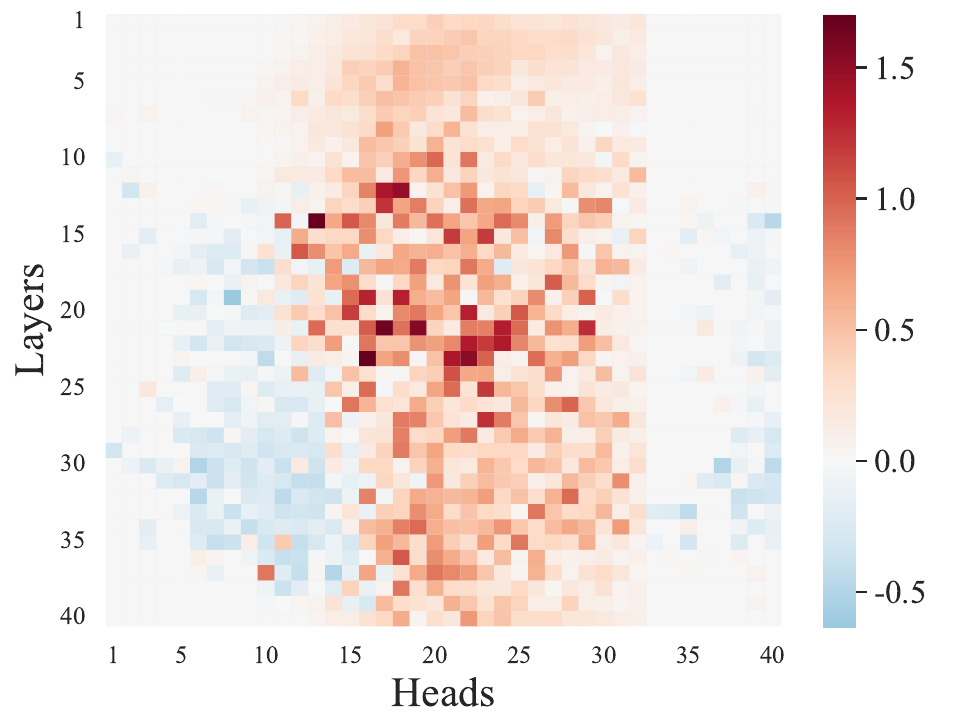}
        \caption{CSQA}
    \end{subfigure}
    \\
    \caption{Attention tracing results across different attention heads on Baichuan2-13B.}
    \label{fig:attn_comp_baichuan}
\end{figure*}

\begin{figure*}[htbp] 
    \centering
	\includegraphics[width=0.8\linewidth]{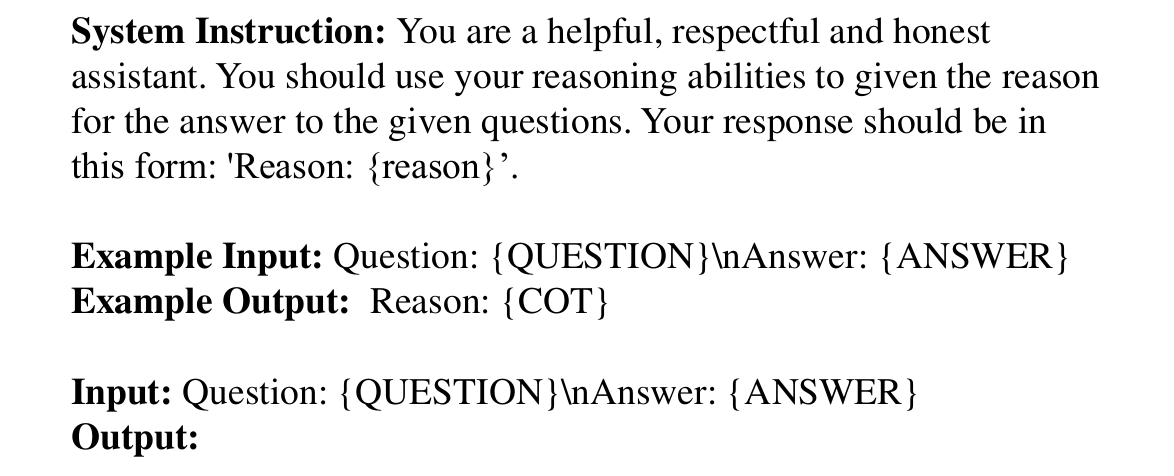}
    \caption{Prompts for correct CoT generation.}
    \label{fig:generate_prompt}
\end{figure*}

Figure \ref{fig:generate_prompt} illustrates the prompt we use for generating correct CoTs from drifting cases.
After the generation, we will manually filter out the wrong CoT and conduct the comparative experiment. We use a 5-shot to generate CoT and concatenate it to the question for our probing experiments.  Figure \ref{fig:attr_comp_csqa} shows the remaining results of this experiment, from which we can get the same conclusions as Section \ref{sec:3.2}. Additionally, we also conduct the supplementary experiments on CSQA and report the results in Figure \ref{fig:attr_comp_csqa_extra_b}.
\paragraph{Attention Tracing Experiment}
Figure \ref{fig:attn_comp_baichuan} reports more results in this experiment.

\section{More Details for Answering Tracing}
\label{sec:inter_trac}
\begin{figure*}[htbp] 
    \centering
    \begin{subfigure}[t]{.49\linewidth}
        \centering
	\includegraphics[width=\linewidth]{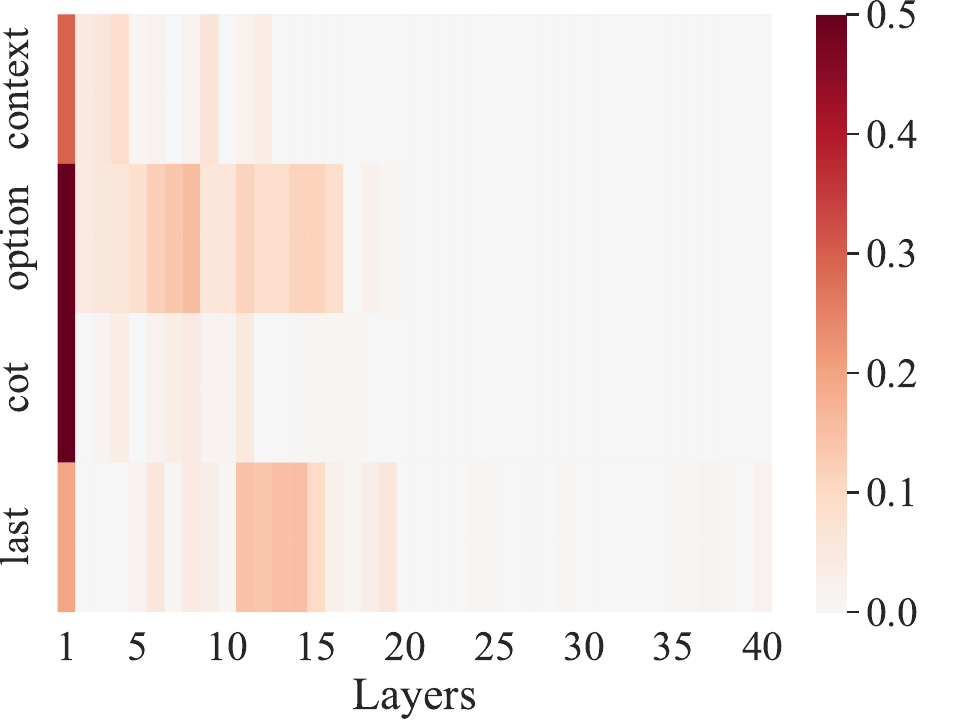}
        \caption{Correct Case's Attn}\label{fig:inter_csqa_attention_cor}
    \end{subfigure}
    \begin{subfigure}[t]{.49\linewidth}
        \centering
	\includegraphics[width=\linewidth]{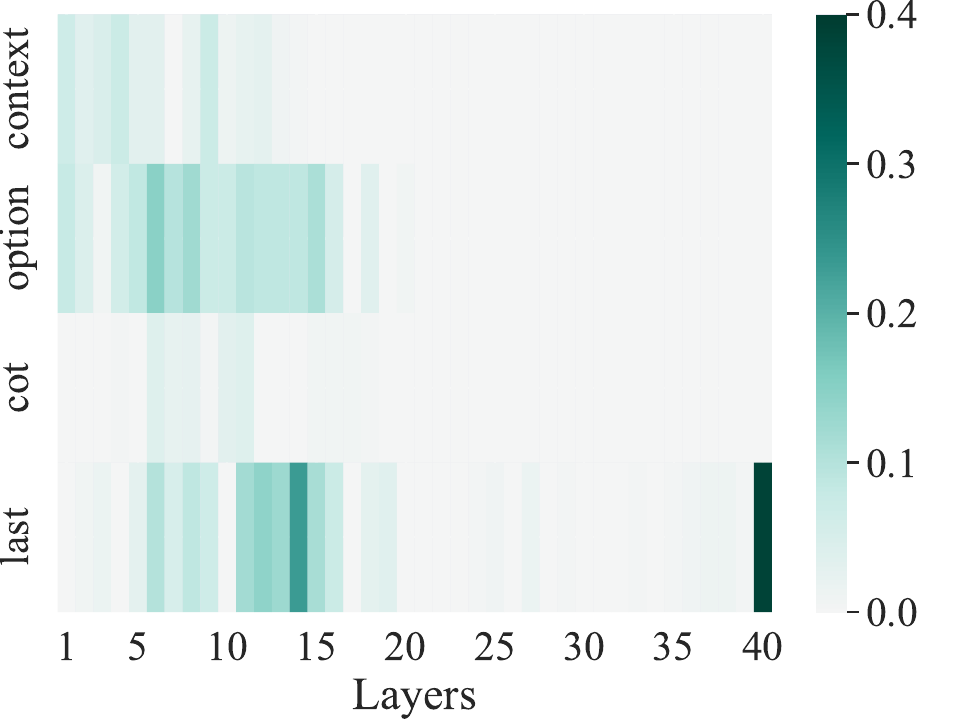}
        \caption{Correct Case's MLP}\label{fig:inter_csqa_mlp_cor}
    \end{subfigure}
     \begin{subfigure}[t]{.49\linewidth}
        \centering
	\includegraphics[width=\linewidth]{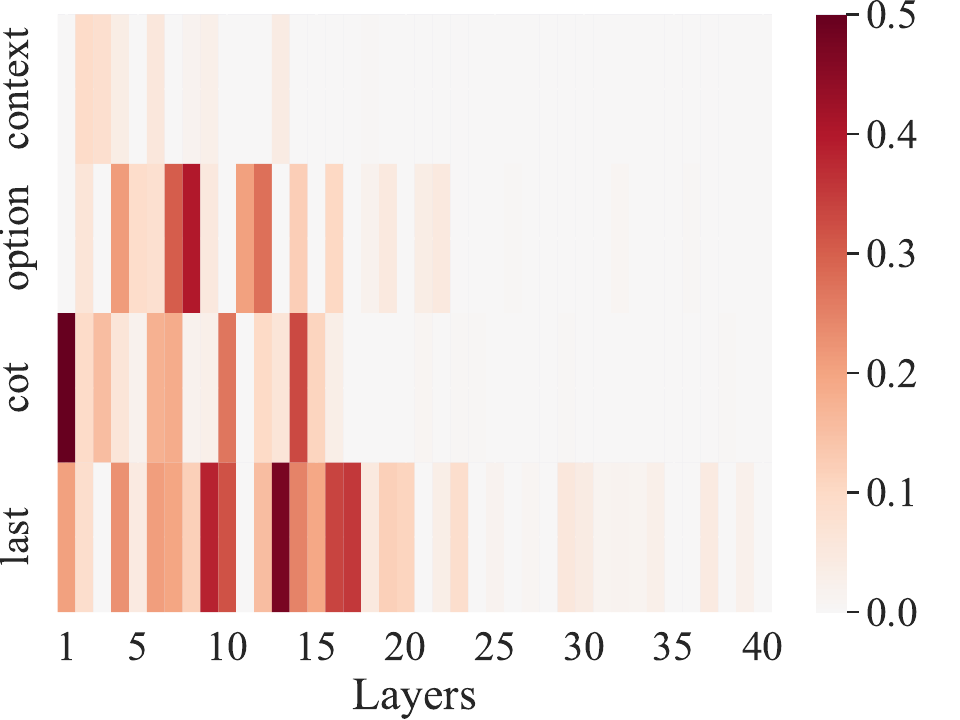}
        \caption{Drifting Case's Attn}\label{fig:inter_csqa_attention_dri}
    \end{subfigure}
    \begin{subfigure}[t]{.49\linewidth}
        \centering
	\includegraphics[width=\linewidth]{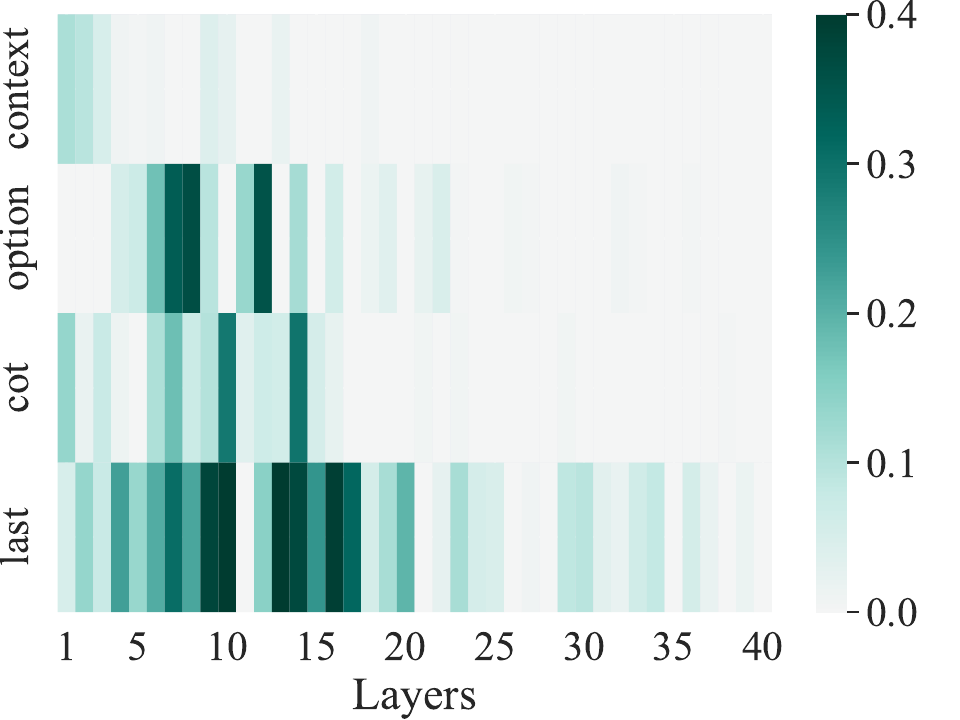}
        \caption{Drifting Case's MLP}\label{fig:inter_csqa_mlp_dri}
    \end{subfigure}
    \\
    \caption{Intervention tracing results on CSQA in correct and drifting answering cases (Llama2-13B).}
    \label{fig:csqa_inter}
\end{figure*}

\begin{figure*}[htbp] 
    \centering
    \begin{subfigure}[t]{.49\linewidth}
        \centering
	\includegraphics[width=\linewidth,  trim=20 0 0 0, clip]{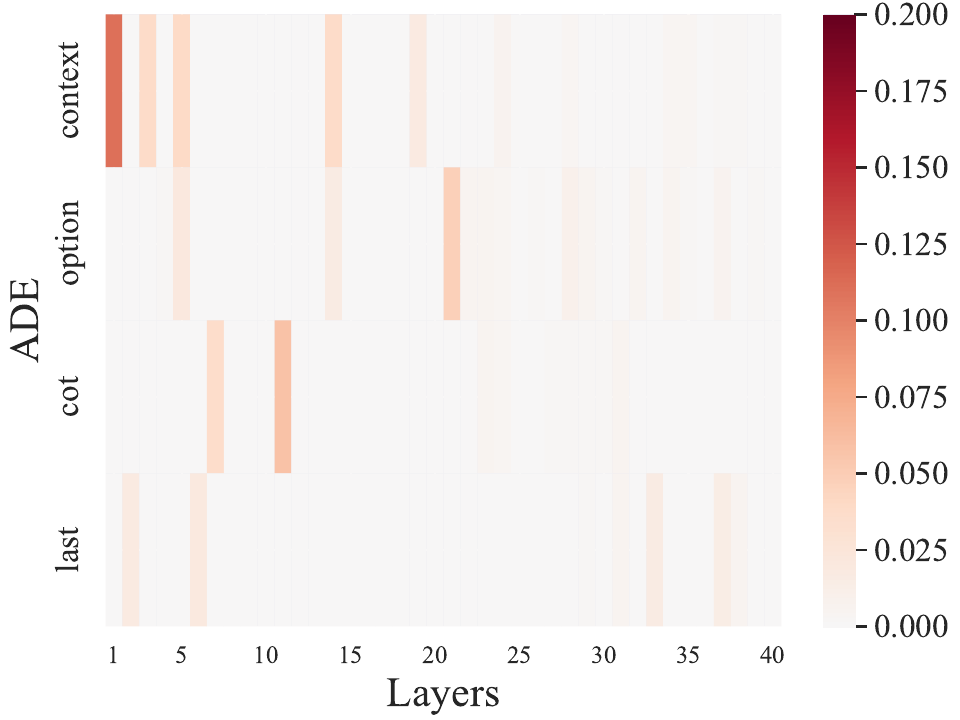}
        \caption{Correct Case's Attn}\label{fig:inter_wino_attention_cor_b}
    \end{subfigure}
    \begin{subfigure}[t]{.49\linewidth}
        \centering
	\includegraphics[width=\linewidth, trim=20 0 0 0, clip]{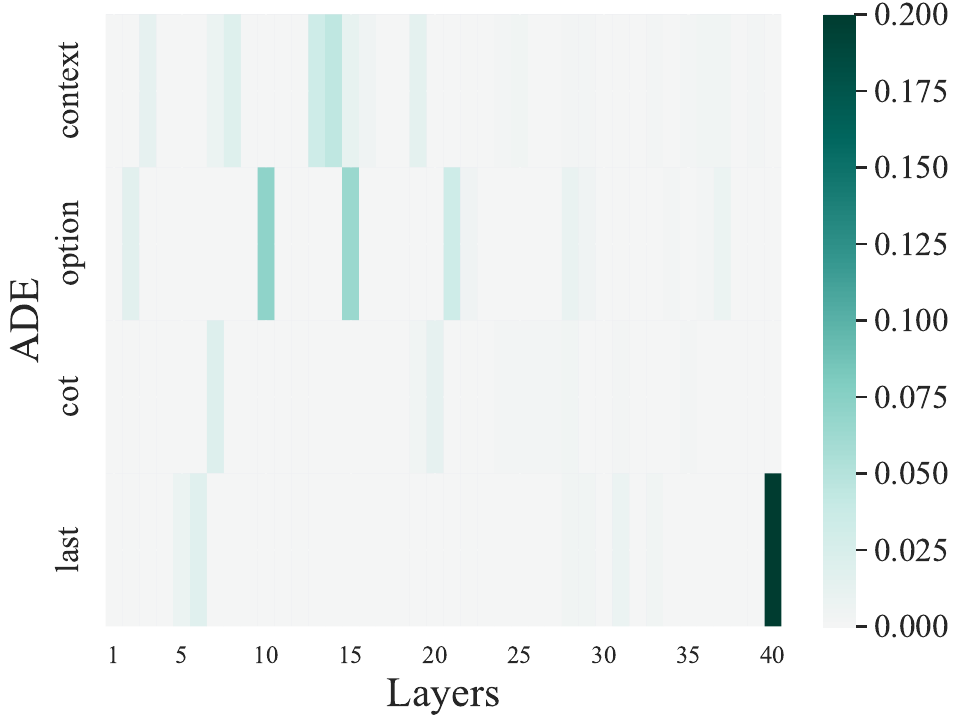}
        \caption{Correct Case's MLP}\label{fig:inter_wino_mlp_cor_b}
    \end{subfigure}
     \begin{subfigure}[t]{.49\linewidth}
        \centering
	\includegraphics[width=\linewidth,  trim=20 0 0 0, clip]{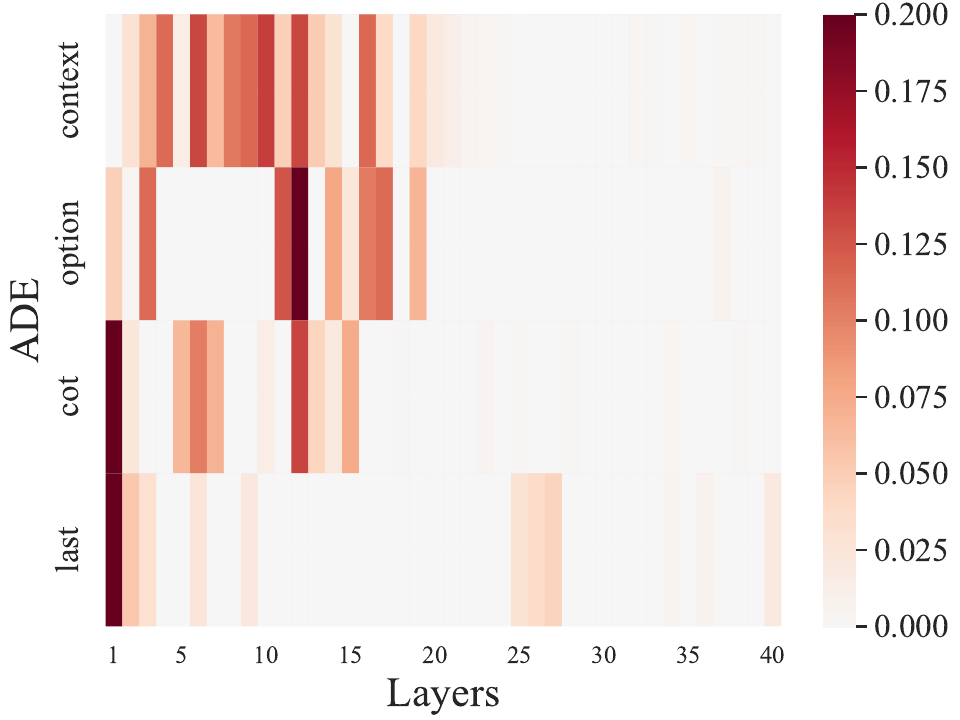}
        \caption{Drifting Case's Attn}\label{fig:inter_wino_attention_dri_b}
    \end{subfigure}
    \begin{subfigure}[t]{.49\linewidth}
        \centering
	\includegraphics[width=\linewidth, trim=20 0 0 0, clip]{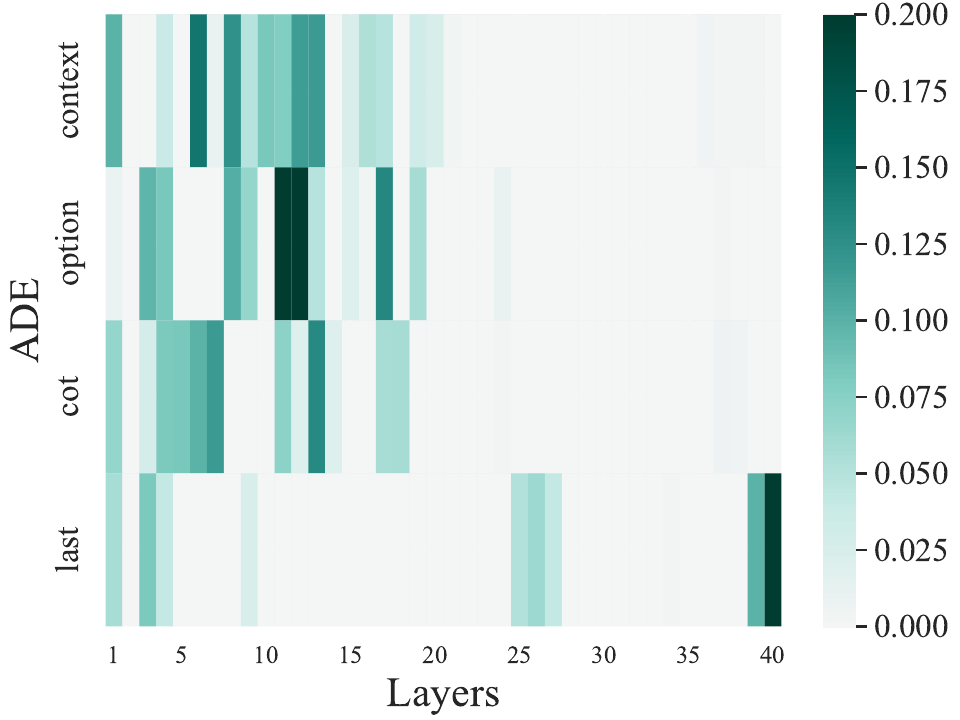}
        \caption{Drifting Case's MLP}\label{fig:inter_wino_mlp_dri_b}
    \end{subfigure}
    \\
    \caption{Intervention tracing results on Winogrande in correct and drifting answering cases (Baichuan2-13B).}
    \label{fig:wino_inter_b}
\end{figure*}

\begin{figure*}[htbp] 
    \centering
    \begin{subfigure}[t]{.49\linewidth}
        \centering
	\includegraphics[width=\linewidth,  trim=20 0 0 0, clip]{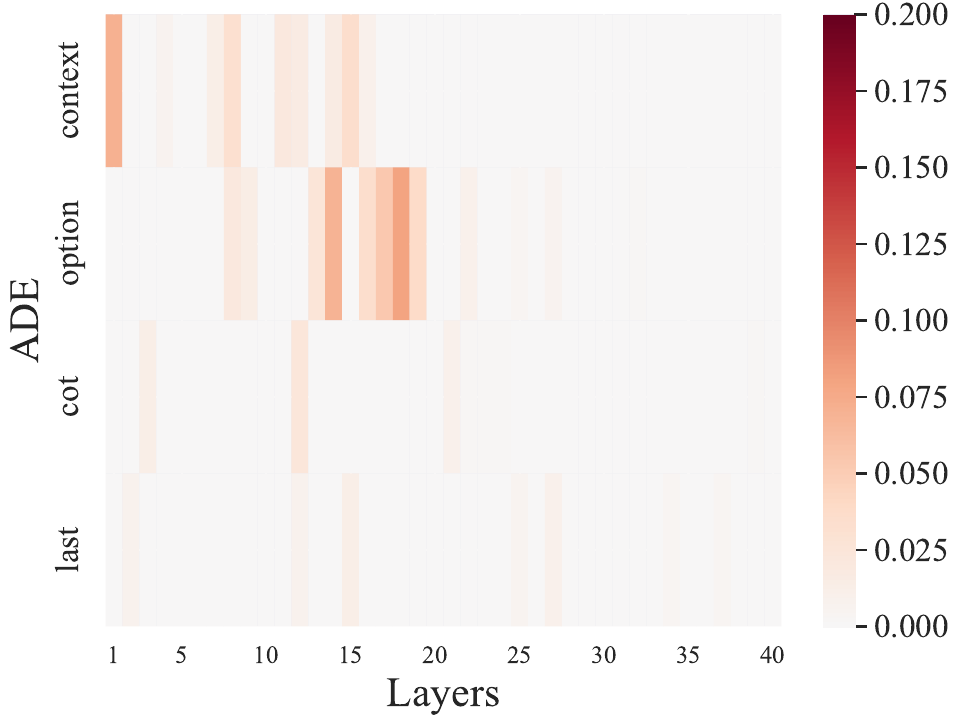}
        \caption{Correct Case's Attn}\label{fig:inter_csqa_attention_cor_b}
    \end{subfigure}
    \begin{subfigure}[t]{.49\linewidth}
        \centering
	\includegraphics[width=\linewidth,  trim=20 0 0 0, clip]{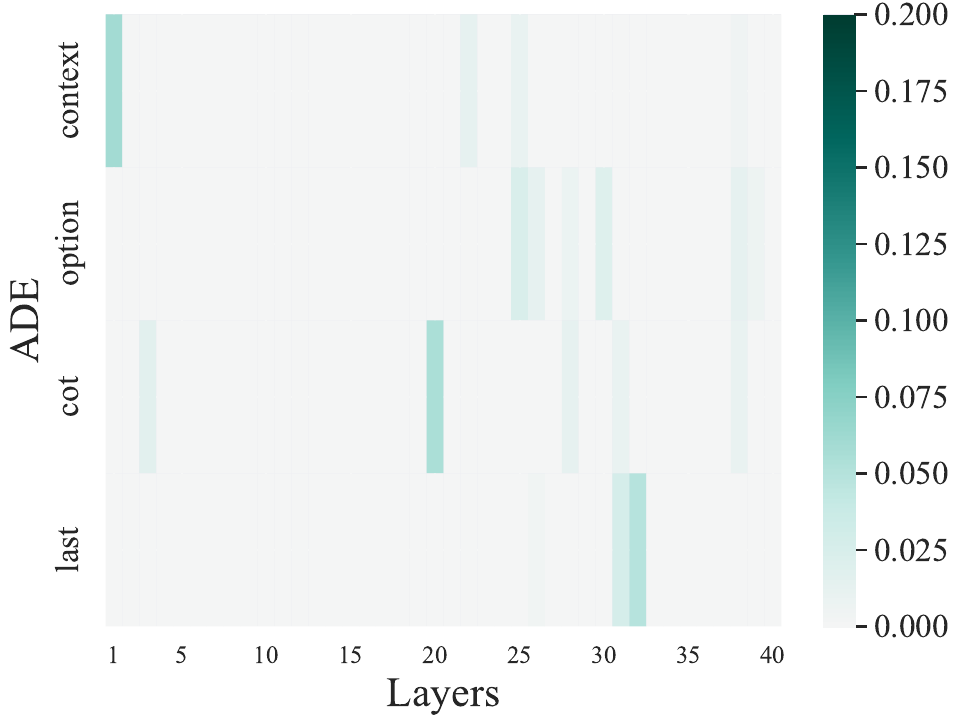}
        \caption{Correct Case's MLP}\label{fig:inter_csqa_mlp_cor_b}
    \end{subfigure}
     \begin{subfigure}[t]{.49\linewidth}
        \centering
	\includegraphics[width=\linewidth,  trim=20 0 0 0, clip]{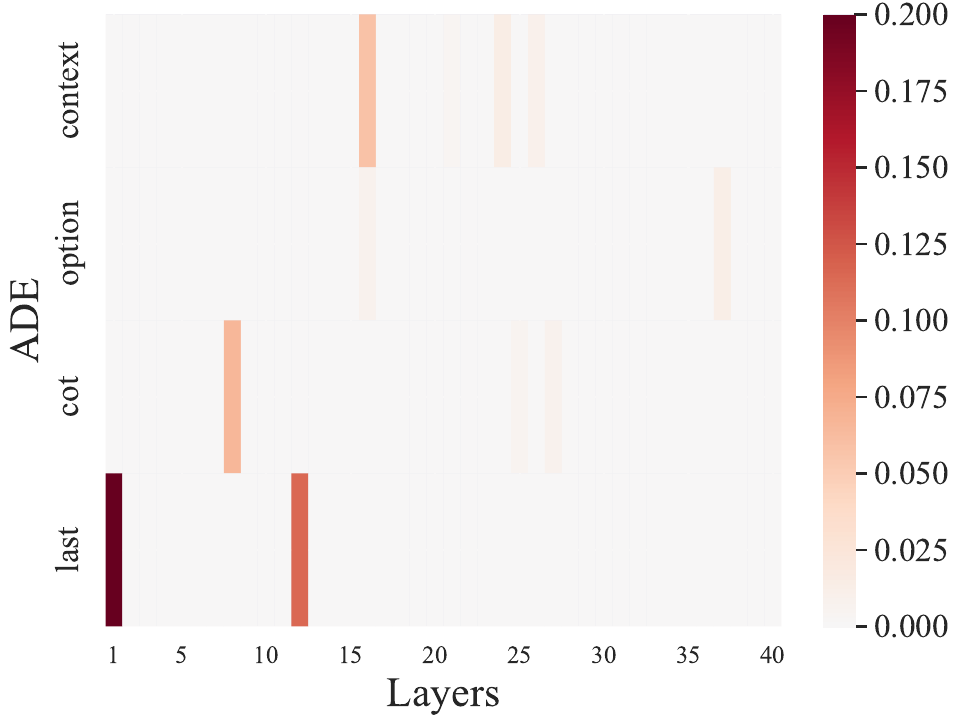}
        \caption{Drifting Case's Attn}\label{fig:inter_csqa_attention_dri_b}
    \end{subfigure}
    \begin{subfigure}[t]{.49\linewidth}
        \centering
	\includegraphics[width=\linewidth,  trim=20 0 0 0, clip]{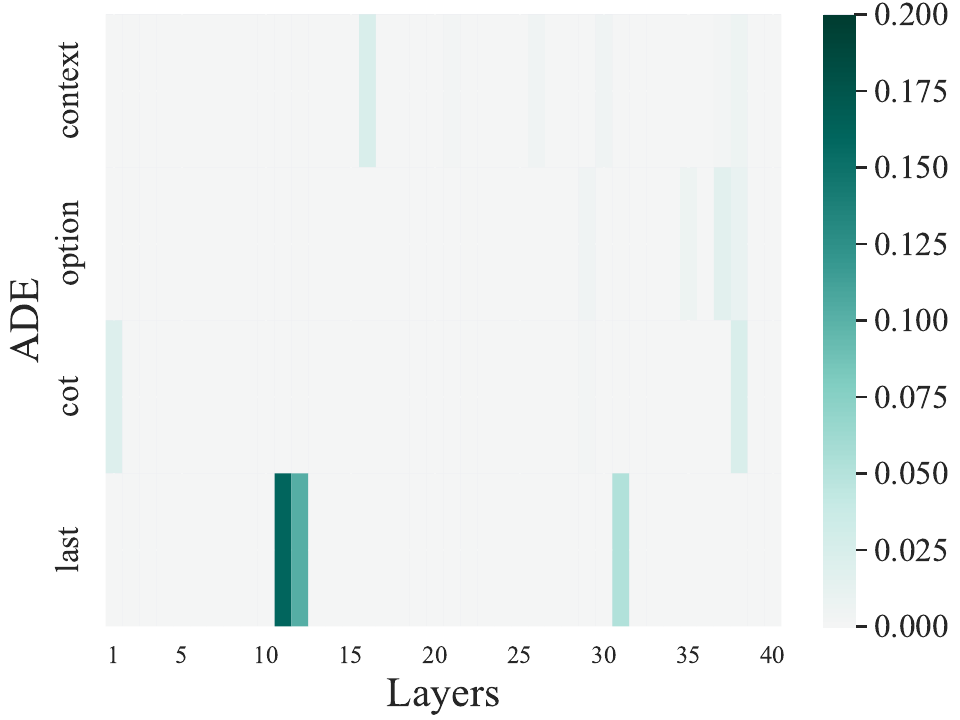}
        \caption{Drifting Case's MLP}\label{fig:inter_csqa_mlp_dri_b}
    \end{subfigure}
    \\
    \caption{Intervention tracing results on CSQA in correct and drifting answering cases (Baichuan2-13B).}
    \label{fig:csqa_inter_b}
\end{figure*}

\begin{figure*}[htbp] 
    \centering
    \begin{subfigure}[t]{.49\linewidth}
        \centering
	\includegraphics[width=\linewidth]{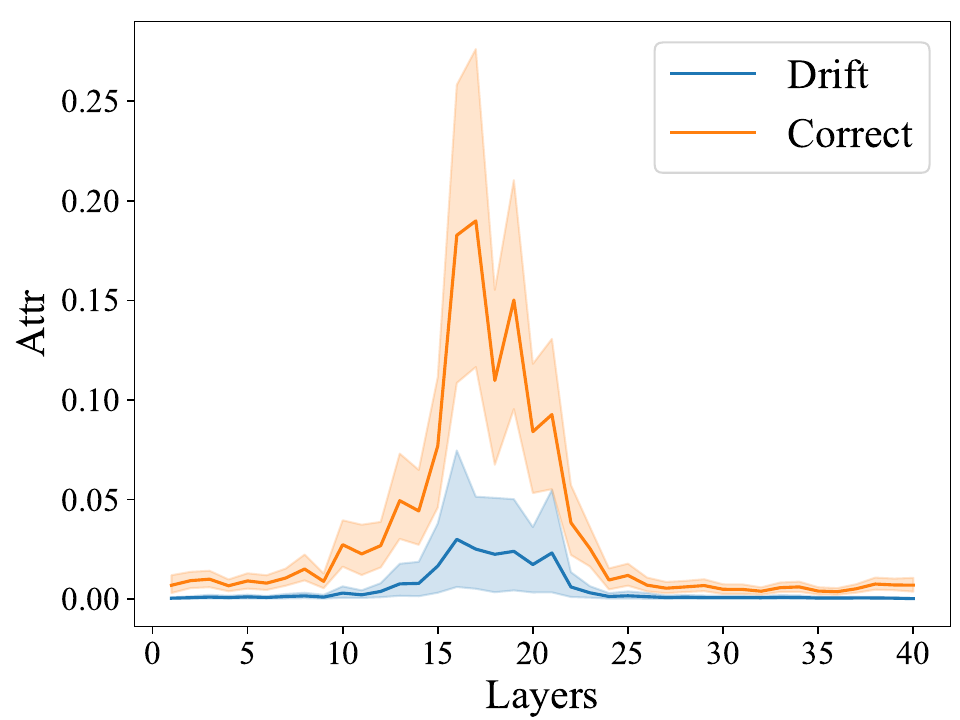}
        \caption{Winogrande}
    \end{subfigure}
    \begin{subfigure}[t]{.49\linewidth}
        \centering
	\includegraphics[width=\linewidth]{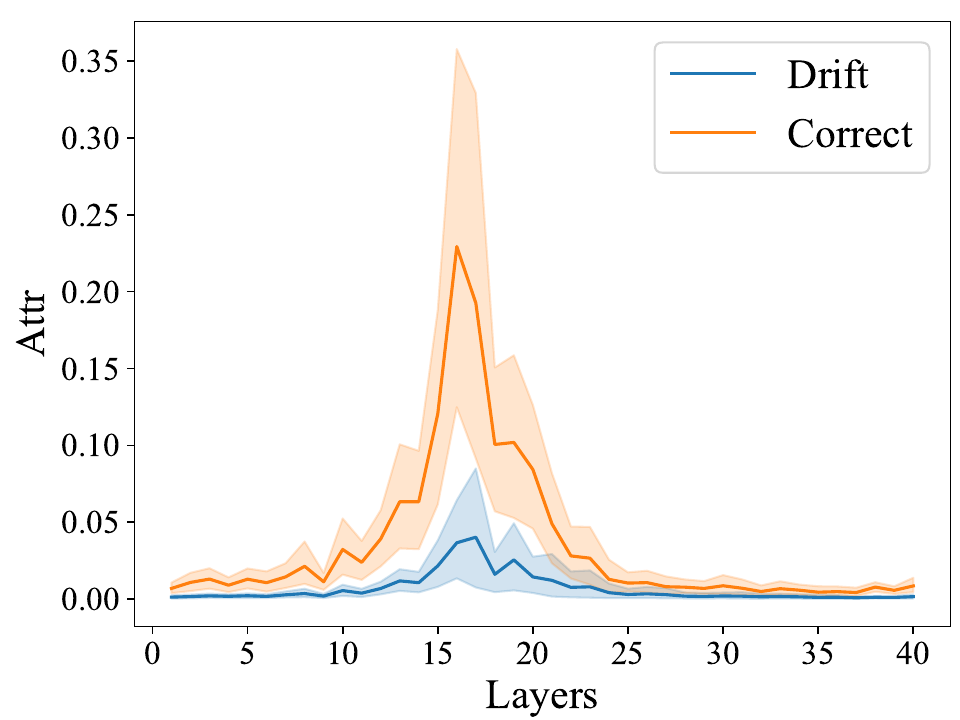}
        \caption{CSQA}
    \end{subfigure}
    \\
    \caption{Attribution tracing results on Baichuan2-13B during the answer generation stage.}
    \label{fig:attr_label_b}
\end{figure*}

\paragraph{Intervention Tracing Experiment}
Figure \ref{fig:csqa_inter}, \ref{fig:wino_inter_b}, \ref{fig:csqa_inter_b} and \ref{fig:attr_label_b} report the remaining intervention tracing experiment results, which are consistent with our conclusion in the main text.

\section{Mitigation Method Implementation}
\label{sec:main_method}
\paragraph{Residual Decoding}
Here, we provide a detailed explanation of Algorithm \ref{alg:RD}. At the beginning, we set the input to the entire question (contexts + options). In line 2, we get the logits from the output of the \textit{LlamaForCausalLM}. In line 6, we calculate the attention score by summing the values on the attention matrix corresponding to the tokens. We use the output character ``\textit{</s>}'' as the termination condition for Llama2-13B generation in line 12.
\paragraph{Serial-Position Swap}
In this method, we swap the positions of the question and the generated CoT, outputting the option with the highest logits score. This method can be implemented under both few-shot and zero-shot settings, demonstrating its cost-efficiency.

\section{More Details for the Mitigation Experiment}
\label{sec:main_exp}
\begin{figure*}[htbp] 
    \centering
	\includegraphics[width=0.8\linewidth]{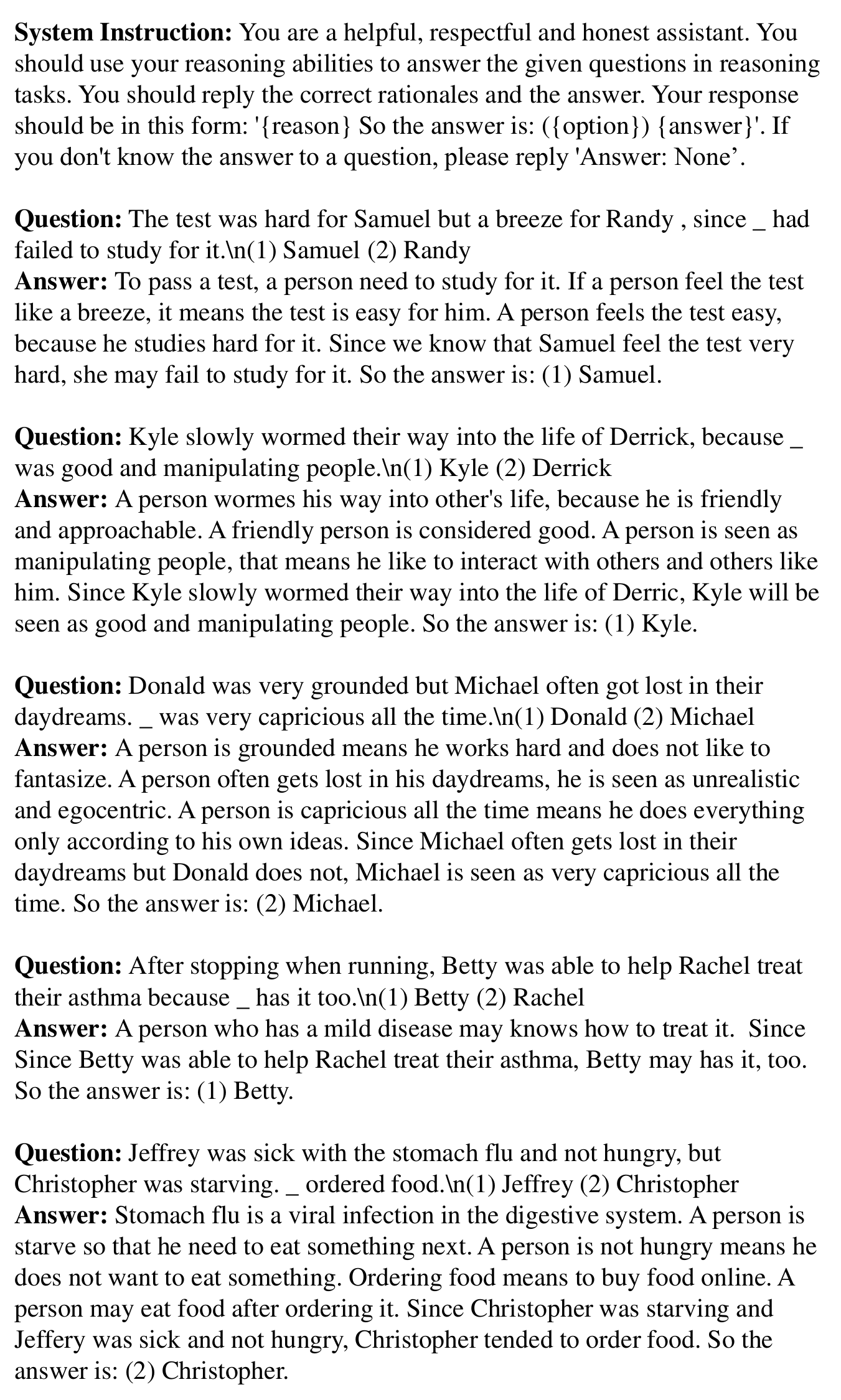}
    \caption{5-shot prompts for Winogrande.}
    \label{fig:cot_wino_prompt}
\end{figure*}

\begin{figure*}[htbp] 
    \centering
	\includegraphics[width=0.8\linewidth]{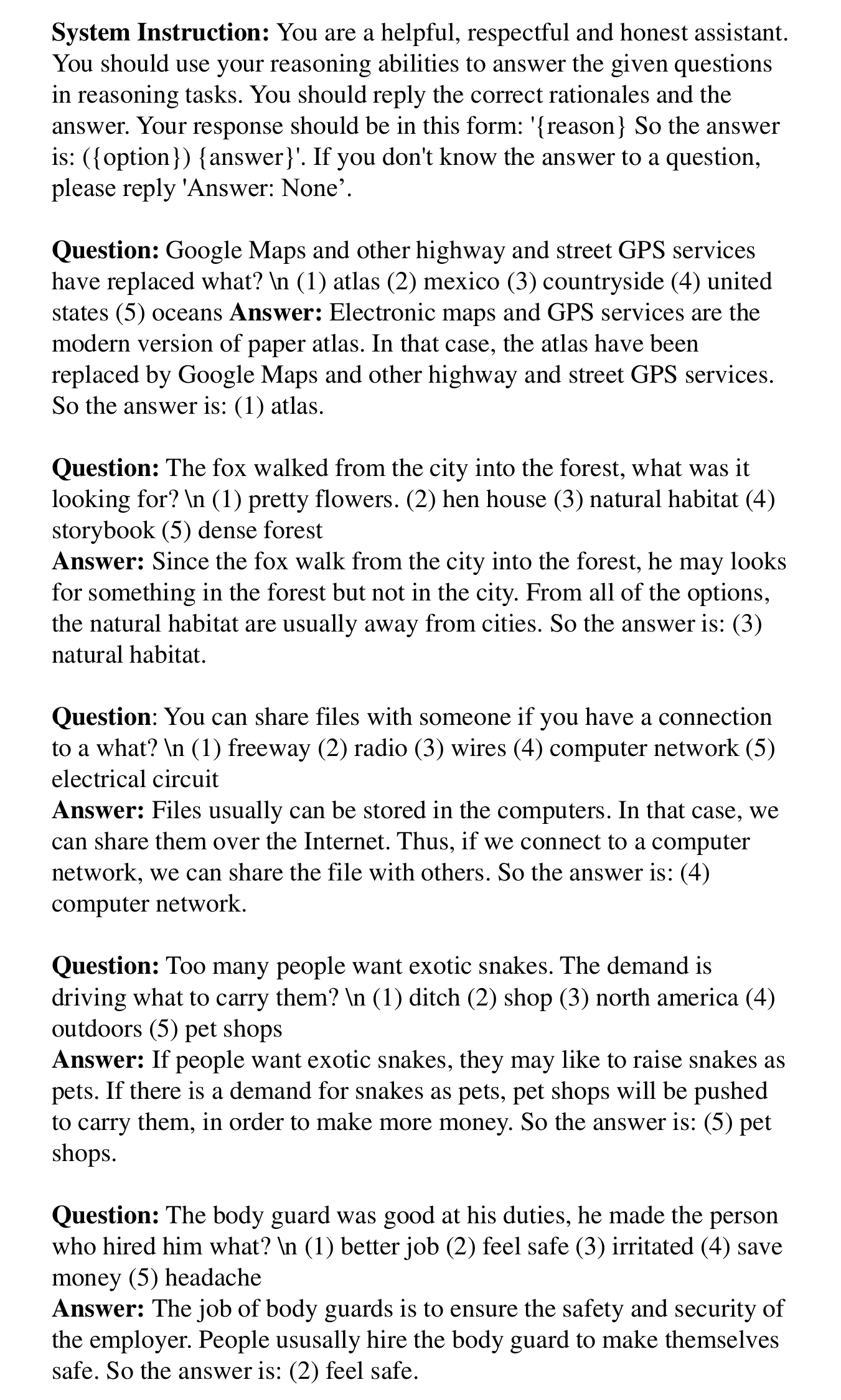}
    \caption{5-shot prompts for CSQA.}
    \label{fig:cot_csqa_prompt}
\end{figure*}

\begin{figure*}[htbp] 
    \centering
	\includegraphics[width=0.9\linewidth]{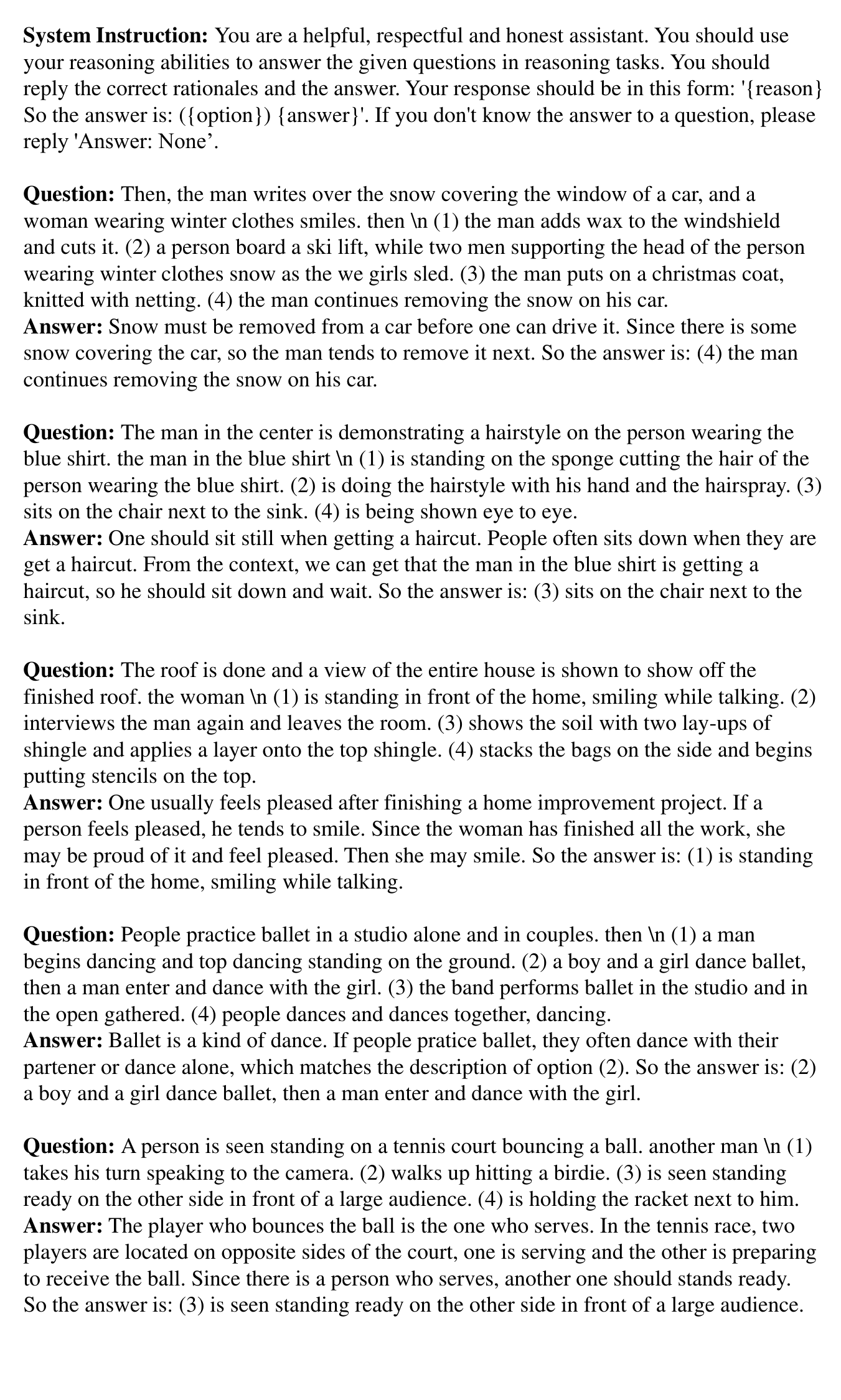}
    \caption{5-shot prompts for HellaSwag.}
    \label{fig:cot_hella_prompt}
\end{figure*}

\begin{figure*}[htbp] 
    \centering
	\includegraphics[width=0.8\linewidth]{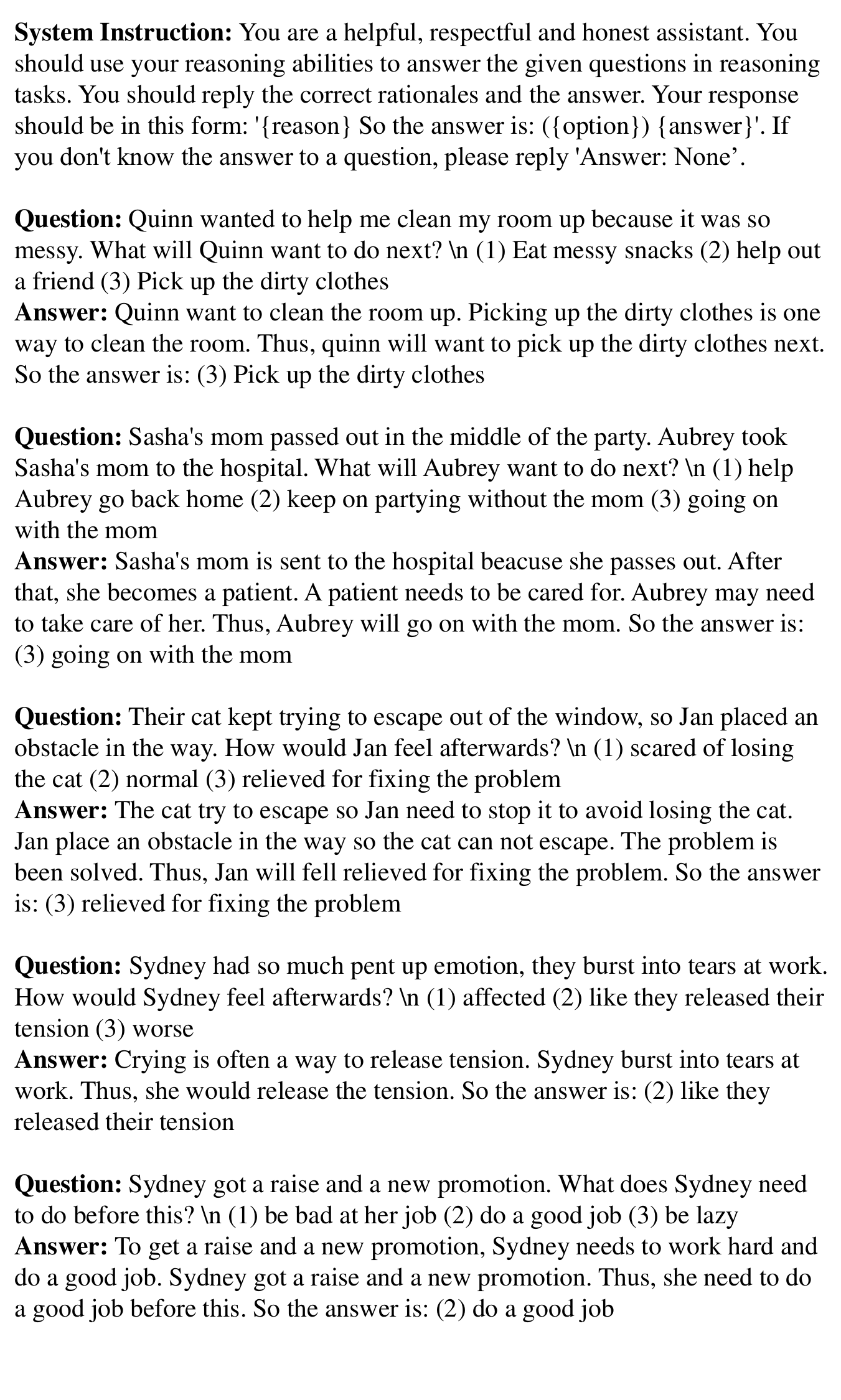}
    \caption{5-shot prompts for SIQA.}
    \label{fig:cot_siqa_prompt}
\end{figure*}

\begin{figure*}[htbp] 
    \centering
	\includegraphics[width=0.8\linewidth]{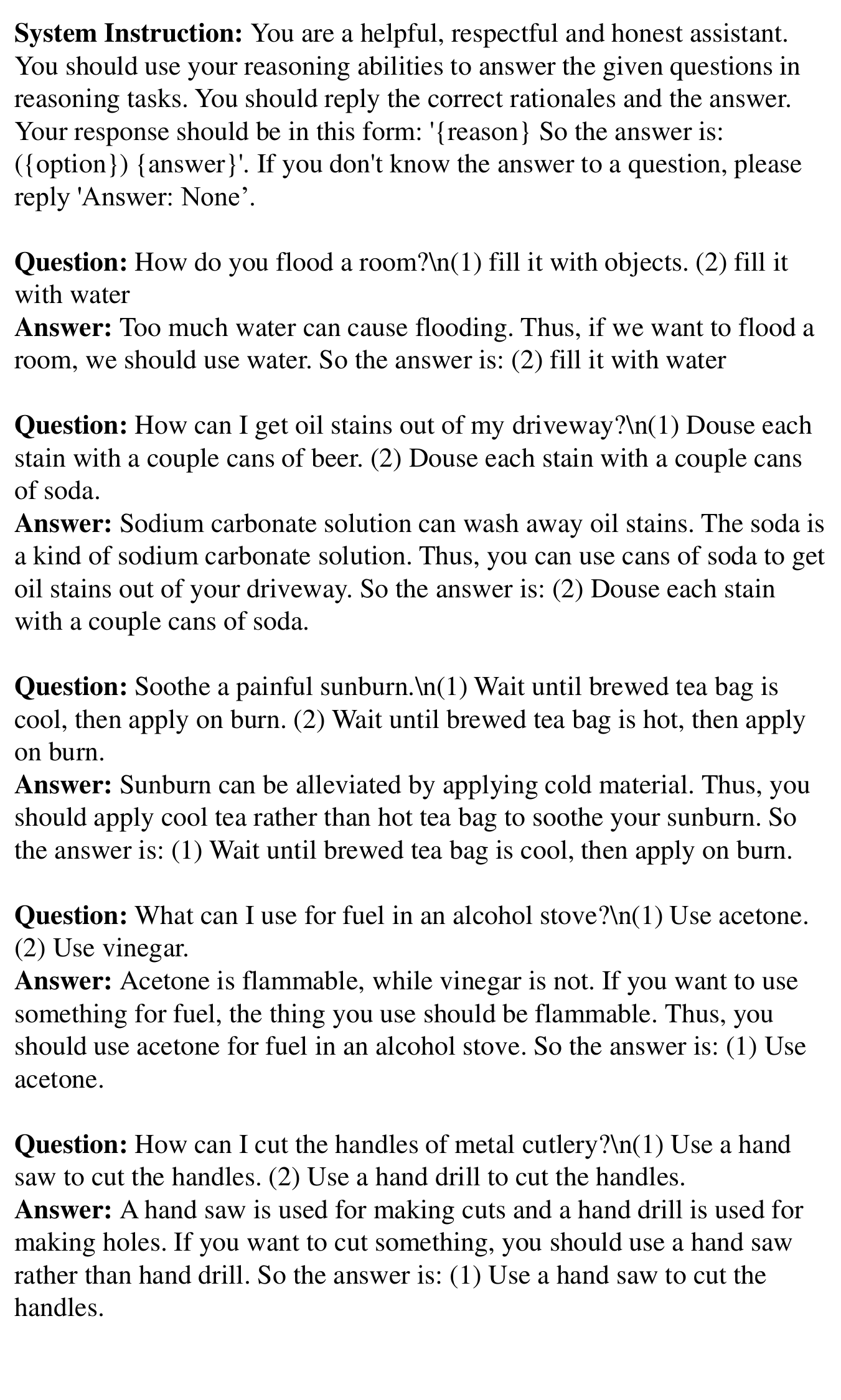}
    \caption{5-shot prompts for PIQA.}
    \label{fig:cot_piqa_prompt}
\end{figure*}

\paragraph{Dataset}
In this experiment, the specific information of all datasets can be found in Table \ref{tab:data_stat}.
\paragraph{Baselines}
For the Self-Consistency method, we sample 5 CoTs and use a majority voting method to select the final predicted answer. For the Self-Refine method, we first conduct one round of CoT reasoning and then follow it with one round of feedback to generate the final answer. For all the baselines, we release the prompts in our source code. We implement all of the methods on \texttt{Llama2-13B-Chat-hf} and \texttt{Baichuan2-13B-Chat}.
\paragraph{Our methods}
In our RD method, we set two hyperparameters --- candidate\_num $n$ and weight $\omega$, and here are their specific values in the experiments: for Winogrande, we set $n$ to $4$ and $\omega$ to $80$, for CSQA, we set $n$ to $10$ and $\omega$ to $135$, for HellaSwag, we set $n$ to $3$ and $\omega$ to $80$, for SIQA, we set $n$ to $10$ and $\omega$ to $160$, for PIQA, we set $n$ to $4$ and $\omega$ to $120$. Additionally, in Figure \ref{fig:cot_wino_prompt},\ref{fig:cot_csqa_prompt},\ref{fig:cot_hella_prompt},\ref{fig:cot_siqa_prompt},\ref{fig:cot_piqa_prompt}, we list our method's few-shot prompts on five datasets. Note that both CoT prompting and our two methods utilize the same prompt.

\begin{table*}[htbp]
\centering
 \scalebox{0.8}{
\begin{tabular}{llcccccccccccc}
\toprule
\multicolumn{2}{l}{\multirow{2}{*}{\textbf{Method}}} & \multicolumn{2}{c}{\textbf{Winogrande}} & \multicolumn{2}{c}{\textbf{CSQA}} & \multicolumn{2}{c}{\textbf{HellaSwag}} & \multicolumn{2}{c}{\textbf{SIQA}} & \multicolumn{2}{c}{\textbf{PIQA}} & \multicolumn{2}{c}{\textbf{Avg}} \\
\cline{3-14}
& & ACC$\uparrow$ & TR$\downarrow$ & ACC$\uparrow$ & TR$\downarrow$& ACC$\uparrow$ & TR$\downarrow$ & ACC$\uparrow$ & TR$\downarrow$& ACC$\uparrow$ & TR$\downarrow$ & ACC$\uparrow$ & TR$\downarrow$  \\
\midrule
\multicolumn{2}{l}{Few-shot Answer} & 59.7  & - & 69.9 & - & \textbf{47.1} & - & 68.8 & - & 60.7 & - & 61.2 & -\\
\multicolumn{2}{l}{Chain-of-Thought} & 58.7  & 34.8 & 66.8 & 37.8 & 41.1 & 27.5 & 66.5 & 37.5 & 66.1 & 26.4 & 59.8 & 32.8\\
\multicolumn{2}{l}{Self-Consistency} & 56.1 & 38.8 & 64.3 & 41.5 & 39.4 & 30.6 & 67.8 & 33.5 & \textbf{70.0} & 21.0 & 59.5 & 33.1\\ 
\multicolumn{2}{l}{Self-Refine} & 59.6 & 35.7 & 66.7 & 38.8 & 39.2 & 29.9 & 62.2 & 43.1 & 62.6 & 31.0 & 58.1 & 35.7\\
\multicolumn{2}{l}{Least-to-Most} & 59.7 & 35.6 & 62.8 & 45.6 & 37.8 & 32.1 & 66.5 & 36.8 & 62.8 & 20.6 & 57.9 & 34.1\\
\multicolumn{2}{l}{Contrasive CoT} & 55.2 & 40.6 & 67.0 & 37.4 & 38.7 & 29.7 & 65.0 & 37.3 & 65.9 & 26.2 & 58.4 & 34.2\\ 
\midrule
\multirow{3}{*}{\textbf{Ours}} &RD Only & 59.2 & 21.1 & 69.5 & 25.3& 44.4 & 19.3 & 68.9 & 25.0 & 67.0 & 12.9 & 61.8 & 20.7\\
&SPS Only &59.9 & 22.8 & 69.7 & 23.5 & 46.3 & 15.2 & \textbf{69.2} & 27.8 & 67.7 & 14.2 & 62.6 & 20.7\\
&RIDERS & \textbf{60.1} & \textbf{18.6}& \textbf{71.3} & \textbf{14.6} & 46.3 & \textbf{13.0} &  69.0 & \textbf{20.5} & 67.2 & \textbf{8.3} & \textbf{62.8} & \textbf{15.0}\\ 
\bottomrule
\end{tabular}}
\caption{Performance comparison across five commonsense reasoning datasets on Baichuan2-13B.}
\label{tab:main_b}
\end{table*}

\begin{table*}[tbp]
\centering

\begin{tabular}{llcccc}
\toprule
\multirow{2}{*}{\textbf{Model}} & \multirow{2}{*}{\textbf{Method}} & \multicolumn{2}{c}{\textbf{Winogrande}} & \multicolumn{2}{c}{\textbf{SIQA}}\\
\cline{3-6}
& & ACC$\uparrow$ & TR$\downarrow$ & ACC$\uparrow$ & TR$\downarrow$  \\
\midrule
\multirow{2}{*}{Mistral-7B} & CoT & 59.2  & 27.9 & 70.2 & 28.2  \\
& RIDERS & \textbf{62.0} & \textbf{15.3}& \textbf{70.6} & \textbf{24.5}\\ 
\midrule
\multirow{2}{*}{GPT-3.5} & CoT & 64.6  & 48.8 & 70.6 & 30.3  \\
& RIDERS & \textbf{70.1} & \textbf{41.7}& \textbf{72.1} & \textbf{29.0}\\ 
\bottomrule
\end{tabular}
\caption{Performance comparison on Mistral-7B and GPT-3.5.}
\label{tab:other_model}
\end{table*}

\paragraph{Results}
In our paper, we choose Llama2-13B for experiments because it is an open-source LLM of suitable size and has great influence. Here, we also repeat all experiments on Baichuan2-13B to verify the generality of our work (see Table \ref{tab:main_b}). Additionally, in Table \ref{tab:other_model}, we show our experimental results on two other representative models: \texttt{Mistral-7B-Instruct-v0.2} and \texttt{gpt-3.5-turbo-1106} (since it is a closed source model, here we only apply the SPS method).

\begin{table*}[tbp]
\centering
\begin{tabular}{lcccccc}
\toprule
\textbf{Strategy} & \textbf{Wino} & \textbf{CSQA} & \textbf{Hella} & \textbf{SIQA} & \textbf{PIQA} & \textbf{Avg}\\
\midrule
Greedy & 5.2 & 2.8 & 4.9 & 3.6 & 5.8 & 4.5\\
Beam & 8.1 & 4.9 & 10.1 & 5.8 & 9.4 & 7.7\\
Ours & 13.4 & 6.6 & 16.2 & 7.4 & 15.9 & 11.9\\
\bottomrule
\end{tabular}
\caption{Decoding time per example comparison.}
\label{tab:time}
\end{table*}

\section{More Details for Cost Analysis}
\label{sec:cost}
Table \ref{tab:time} illustrates the time cost comparison of our residual decoding methods with other strategies. Here we set \textit{num\_beams} in the beam search strategy as $5$, and \textit{candidate\_num} in the RD strategy as $3$. We compute the average seconds cost per example over 50 samples for each dataset. On average, our decoding strategy takes 2.6 times longer than greedy search and 1.4 times longer than beam search. This reflects that our decoding method has significantly stronger performance while having a comparable overall time to these main decoding strategies.

\newpage

\begin{figure*}[htbp] 
    \centering
    \begin{subfigure}[t]{.49\linewidth}
        \centering
	\includegraphics[width=\linewidth]{wino_attr_comparison_label_method.pdf}
        \caption{Winogrande}
    \end{subfigure}
    \begin{subfigure}[t]{.49\linewidth}
        \centering
	\includegraphics[width=\linewidth]{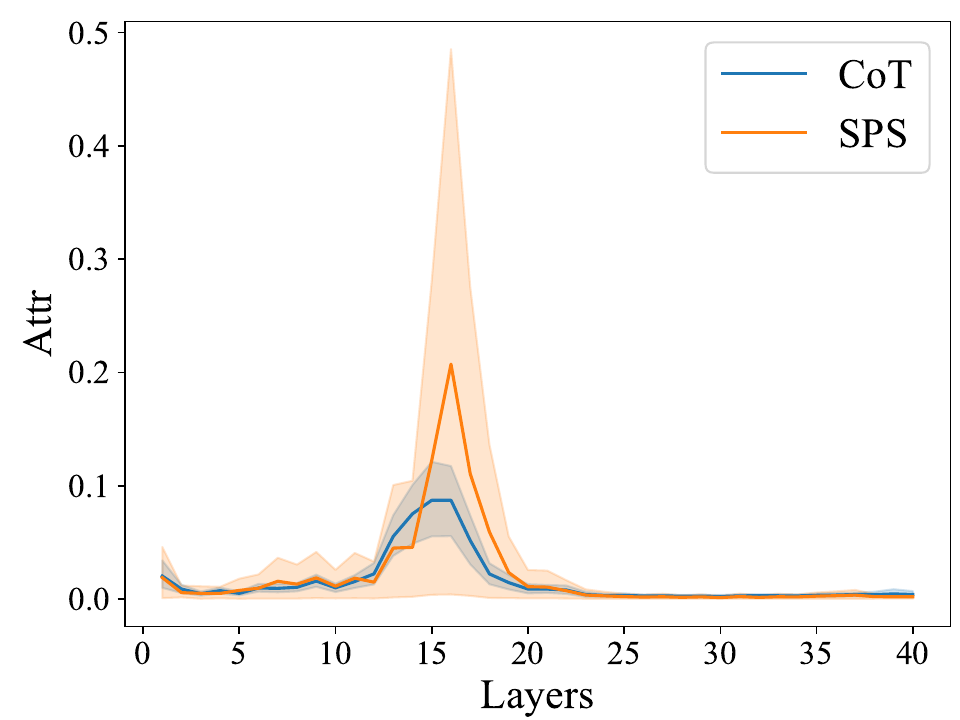}
        \caption{CSQA}
    \end{subfigure}
    \\
    \caption{Information flow comparison on CSQA after applying our two methods.}
    \label{fig:attr_method_csqa}
\end{figure*}

\end{document}